\documentclass[journal]{IEEEtran}

\usepackage{cite}
\usepackage{pifont}
\usepackage{xcolor}
\usepackage{amssymb}
\usepackage{acronym}
\usepackage{graphicx}
\usepackage{textcomp}
\usepackage{hyperref}
\usepackage{multirow}
\usepackage{booktabs}
\usepackage{subcaption}
\usepackage{algorithmic}
\usepackage{amsmath,amssymb,amsfonts}

\acrodef{IR}{Infrared}
\acrodef{FIR}{Far-Infrared}
\acrodef{CV}{Color Visible}
\acrodef{FoV}{Field-of-View}
\acrodef{NIR}{Near-Infrared}
\acrodef{VIS}{Visual-Optical}
\acrodef{MIR}{Middle-Infrared}
\acrodef{TIR}{Thermal-Infrared}
\acrodef{GV}{Grayscale Visible}
\acrodef{LWIR}{Long-wave Infrared}
\acrodef{SWIR}{Short-wave Infrared}
\acrodef{CAN bus}{Controller Area Network}
\acrodef{LiDAR}{Light Detection And Ranging}

\acrodef{MR}{Miss Rate}
\acrodef{FP}{False Positive}
\acrodef{fps}{frames per second}
\acrodef{ROI}{Region of Interest}
\acrodef{IoU}{Intersection over Union}
\acrodef{FLOPS}{Floating-Point Operations per Second}
\acrodef{mAP}{mean Average Precision}
\acrodef{AP}{Average Precision}

\acrodef{SK}{Select Kernel}
\acrodef{I2I}{Image-to-Image}
\acrodef{TV}{Total Variation}
\acrodef{NIN}{Network-In-Network}
\acrodef{ViT}{Vision Transformer}
\acrodef{YOLO}{You Only Look Once}
\acrodef{DNN}{Deep Neural Network}
\acrodef{LS-GAN}{Least-Squares GAN}
\acrodef{MD}{Modality Distillation}
\acrodef{KD}{Knowledge Distillation}
\acrodef{HE}{Histogram Equalization}
\acrodef{AR-CNN}{Aligned Region CNN}
\acrodef{FWN}{Fusion Weight Network}
\acrodef{BDT}{Boosted Decision Tree}
\acrodef{SVM}{Support Vector Machine}
\acrodef{CaRF}{Cascade Random Forest}
\acrodef{GAT}{Graph-Attention Network}
\acrodef{FPN}{Feature Pyramid Network}
\acrodef{RPN}{Region Proposal Network}
\acrodef{RFA}{Region Feature Alignment}
\acrodef{RNN}{Recurrent Neural Network}
\acrodef{SVR}{Support Vector Regression}
\acrodef{F-MSE}{Focal Mean Square Error}
\acrodef{DCT}{Discrete Cosine Transform}
\acrodef{SKNet}{Selective Kernel Network}
\acrodef{TE-GAN}{Thermal-Enhancement GAN}
\acrodef{VST}{Visual Salient Transformer}
\acrodef{ProbEn}{Probabilistic Ensembling}
\acrodef{GFD-SSD}{Gated Fusion Double SSD}
\acrodef{DDFF}{Double Deep Feature Fusion}
\acrodef{MB-Net}{Modality Balance Network}
\acrodef{ResNet-152}{Residual Network-152}
\acrodef{MFA}{Multi-scale Fusion Attention}
\acrodef{CFL}{Cross-modal Feature Learning}
\acrodef{Fast R-CNN}{Fast Region-based CNN}
\acrodef{CNN}{Convolutional Neural Network}
\acrodef{SSD}{Single Shot MultiBox Detector}
\acrodef{RRN}{Region Reconstruction Network}
\acrodef{CAM}{Channel-wise Attention Module}
\acrodef{SAM}{Spatial-wise Attention Module}
\acrodef{MSDN}{Multi-Scale Detection Network}
\acrodef{GAN}{Generative Adversarial Network}
\acrodef{UDA}{Unsupervised Domain Adaptation}
\acrodef{MuFEm}{Multi-modal Feature Embedding}
\acrodef{HOG}{Histogram of Oriented Gradients}
\acrodef{AHE}{Adaptive Histogram Equalization}
\acrodef{GRPN}{Ground-Region Proposal Network}
\acrodef{MANN}{Memory Augmented Neural Network}
\acrodef{SSTN}{Self-Supervised Thermal Network}
\acrodef{GAFF}{Guided Attentive Feature Fusion}
\acrodef{Faster R-CNN}{Faster Region-based CNN}
\acrodef{UFF}{Uncertainty-aware Feature Fusion}
\acrodef{RCAB}{Residual Channel Attention Block}
\acrodef{ARPN}{Automatic Region Proposal Network}
\acrodef{DMAF}{Differential Modality Aware Fusion}
\acrodef{DSMN}{Double-Stream Multispectral Network}
\acrodef{GMA-CycleGAN}{Gray Mask Attention-CycleGAN}
\acrodef{AWGN}{Additive Random White Gaussian Noise}
\acrodef{IAFA}{Illumination Aware Feature Alignment}
\acrodef{UCG}{Uncertainty-aware Cross-modal Guiding}
\acrodef{IAF R-CNN}{Illumination-Aware Faster R-CNN}
\acrodef{TS-RPN}{Two Stream Region Proposal Network}
\acrodef{WDSR}{Wide Activation Deep Super-Resolution}
\acrodef{CMKD}{Cross Modality Knowledge Distillation}
\acrodef{MSFFN}{MultiSpectral Feature Fusion Network}
\acrodef{CSPNet}{Center and Scale Prediction Network}
\acrodef{SCoFA}{Spatio-Contextual Feature Aggregation}
\acrodef{RRRNet}{Recurrent Residual Refinement Network}
\acrodef{HLID}{Histogram of Local Intensity Difference}
\acrodef{HAFNet}{Hierarchical Attentive Fusion Network}
\acrodef{SIDNet}{Scene-Illumination Disentangled Network}
\acrodef{RCCA}{Reference box Constrained Cross-Attention}
\acrodef{R-FCN}{Region-based Fully Convolutional Network}
\acrodef{PiCANet}{Pixel-wise Contextual Attention Network}
\acrodef{RISNet}{Redundant Information Suppression Network}
\acrodef{CIAN}{Cross-Modality Interactive Attention Network}
\acrodef{TCEFNet}{Texture-Contrast Enhancement Fusion Network}
\acrodef{DMFFNet}{Dual-modal Multi-scale Feature Fusion Network}
\acrodef{MLF-FRCNN}{Multi-Layer Fusion network based on Faster R-CNN}
\acrodef{IT-MN}{Illumination and Temperature-aware Multispectral Network}
\acrodef{TIR-ACF}{Thermal Infrared Radiometric Cumulative Channel Feature}
\acrodef{TE-VGAN}{Thermal Enhancement Vision Generative Adversarial Network}
\acrodef{AED-CNN}{Attention-guided Encoder-Decoder Convolutional Neural Network}

\acrodef{GFU}{Gated Fusion Units}
\acrodef{CSM}{Contour Saliency Map}
\acrodef{MSR}{Multi-Spectral Recalling}
\acrodef{FAM}{Feature Attention Module}
\acrodef{CWF}{Channel Weighting Fusion}
\acrodef{HFA}{High-Frequency Assistant}
\acrodef{AFA}{Adjacent Feature Aggregation}
\acrodef{FTM}{Feature Transformation Module}
\acrodef{CBN}{Convolutional Backbone Network}
\acrodef{APF}{Accumulated Probability Fusion}
\acrodef{LPR}{Large-scale Pedestrian Recalling}
\acrodef{MFA}{Multi-modality Feature Alignment}
\acrodef{SPA}{Scale‐aware Permutated Attention}
\acrodef{GAFF}{Guided Attentive Feature Fusion}
\acrodef{CFT}{Cross-modality Fusion Transformer}
\acrodef{ADAS}{Advanced Driver Assistance Systems}
\acrodef{i-IAN}{improved Illumination–Aware Network}
\acrodef{MCFF}{Multi-spectral Channel Feature Fusion}
\acrodef{CIEM}{Cascaded Information Enhancement Module}
\acrodef{CSSA}{Channel Switching and Spatial Attention}
\acrodef{CMDAF}{Cross-Modality Differential Aware Fusion}
\acrodef{CFCM}{Cross-modality Feature Complementary Module}
\acrodef{CAFFM}{Cross-modal Attention Feature Fusion Module}
\acrodef{ASG-LPF}{Adaptive Soft-Gated Light Perception Fusion}
\acrodef{MFEV}{Multiscale Feature Extraction of Visible images}
\acrodef{HCAF}{Hierarchical Content-dependent Attentive Fusion}
\acrodef{MFEI}{Multiscale Feature Extraction of Infrared images}
\acrodef{MS-DETR}{MultiSpectral Pedestrian DEtection TRansformer}
\acrodef{AFEFM}{Attention-based Feature Enhancement Fusion Module}
\acrodef{OCS-LBPs}{Oriented Center-Symmetric Local Binary Patterns}

\acrodef{KMU}{Keimyung University}
\acrodef{OSU}{Ohio State University}
\acrodef{TIV}{Thermal Infrared Video}
\acrodef{FLIR}{Forward Looking InfraRed}
\acrodef{NTPD}{Night-Time Pedestrian Dataset}
\acrodef{LLVIP}{Low Light Visible Image Person}
\acrodef{SCUT}{South China University of Technology}
\acrodef{ZUT}{Zachodniopomorski Uniwersytet Technologiczny}
\acrodef{KAIST}{Korea Advanced Institute of Science and Technology}
\acrodef{LSI-FIR}{Laboratorio de Sistemas Inteligentes Far-Infrared}
\acrodef{LITIV}{Laboratoire d'Interprétation et de Traitement d'Images et Vidéo}

\newcommand{\wrt}{w.r.t. }
\newcommand{\cmark}{\ding{51}}

\newcommand{\etc}{\textit{etc. }}
\newcommand{\eg}{\textit{e.g., }}
\newcommand{\ie}{\textit{i.e., }}
\newcommand{\etal}{\textit{et al. }}

\begin{document}
\title{Pedestrian Detection in Low-Light Conditions: A Comprehensive Survey}

\author{
    \IEEEauthorblockN{Bahareh Ghari}
    \IEEEauthorblockA{\textit{Department of Computer Engineering} \\
    \textit{University of Guilan} \\
    Rasht, Iran \\
    baharehghari@msc.guilan.ac.ir}
    \and
    \IEEEauthorblockN{Ali Tourani}
    \IEEEauthorblockA{\textit{Interdisciplinary Centre for Security,} \\ \textit{Reliability, and Trust (SnT)} \\
    \textit{University of Luxembourg}\\
    Luxembourg, Luxembourg \\
    ali.tourani@uni.lu}
    \and
    \IEEEauthorblockN{Asadollah Shahbahrami}
    \IEEEauthorblockA{\textit{Department of Computer Engineering} \\
    \textit{University of Guilan} \\
    Rasht, Iran \\
    shahbahrami@guilan.ac.ir}
    \and
    \IEEEauthorblockN{Georgi Gaydadjiev}
    \IEEEauthorblockA{\textit{Faculty of Science and Engineering} \\
    \textit{University of Groningen} \\
    Groningen, Netherlands \\
    g.gaydadjiev@rug.nl}
}

\author{
    Bahareh~Ghari, Ali~Tourani, Asadollah~Shahbahrami, and~Georgi~Gaydadjiev
    \thanks{Bahareh Ghari and Asadollah Shahbahrami are with the Department of Computer Engineering, University of Guilan, Iran.
    \tt{\small{baharehghari@msc.guilan.ac.ir}, \small{shahbahrami@guilan.ac.ir}}    
    }
    \thanks{Ali Tourani is with the Automation and Robotics Research Group, Interdisciplinary Centre for Security, Reliability, and Trust (SnT), University of Luxembourg, Luxembourg. \tt{\small{ali.tourani@uni.lu}}.}
    \thanks{Georgi Gaydadjiev is with the Computer Engineering Laboratory, Delft University of Technology, The Netherlands. \tt{\small{g.n.gaydadjiev@tudelft.nl}}.}
}

\markboth{Journal of \LaTeX\ Class Files,~Vol.~14, No.~8, August~2015}%
{Shell \MakeLowercase{\textit{et al.}}: Bare Demo of IEEEtran.cls for IEEE Journals}

\maketitle

\begin{abstract}
Pedestrian detection remains a critical problem in various domains, such as computer vision, surveillance, and autonomous driving.
In particular, accurate and instant detection of pedestrians in low-light conditions and reduced visibility is of utmost importance for autonomous vehicles to prevent accidents and save lives.
This paper aims to comprehensively survey various pedestrian detection approaches, baselines, and datasets that specifically target low-light conditions.
The survey discusses the challenges faced in detecting pedestrians at night and explores state-of-the-art methodologies proposed in recent years to address this issue.
These methodologies encompass a diverse range, including deep learning-based, feature-based, and hybrid approaches, which have shown promising results in enhancing pedestrian detection performance under challenging lighting conditions.
Furthermore, the paper highlights current research directions in the field and identifies potential solutions that merit further investigation by researchers.
By thoroughly examining pedestrian detection techniques in low-light conditions, this survey seeks to contribute to the advancement of safer and more reliable autonomous driving systems and other applications related to pedestrian safety.
Accordingly, most of the current approaches in the field use deep learning-based image fusion methodologies (\ie early, halfway, and late fusion) for accurate and reliable pedestrian detection.
Moreover, the majority of the works in the field (approximately $48\%$) have been evaluated on the \textit{KAIST} dataset, while the real-world video feeds recorded by authors have been used in less than six percent of the works.
\end{abstract}


\IEEEpeerreviewmaketitle

\section{Introduction}
\label{sec_intro}

\IEEEPARstart{A}{utomatic} identification and localization of pedestrians in images or video frames captured by visual sensors have become increasingly vital in the computer vision domain.
Pedestrian detection has use cases in various fields, such as autonomous vehicles \cite{boukerche2021design, sha2022performance}, surveillance systems \cite{kim2019fast, oluyide2022automatic, oltean2019pedestrian}, and robotics \cite{zou2022active, zhao2022pedestrian, pang2019efficient}.
This task can be challenging to resolve in real-world scenarios, as there are different factors to consider for accurate performance.
Accordingly, various illumination conditions, dissimilar pedestrian appearances and poses, occlusion, camouflage, and cluttered backgrounds can bring about issues for pedestrian detection systems \cite{gawande2020pedestrian}.
Regarding the introduced challenges, low illumination is the leading problem compared to others, as it has a natural or environment-related cause and cannot be prevented or naturally handled.
It can be due to the time of the day, geographical location of where the scene is captured, weather conditions, \etc
For instance, in some Scandinavian countries, especially during winter, the sunrise to sunset time can be less than ten hours and the pedestrian detection systems need to be adapted to the challenging scenarios.

\begin{figure}
    \centering
    \includegraphics[width=1.0\columnwidth]{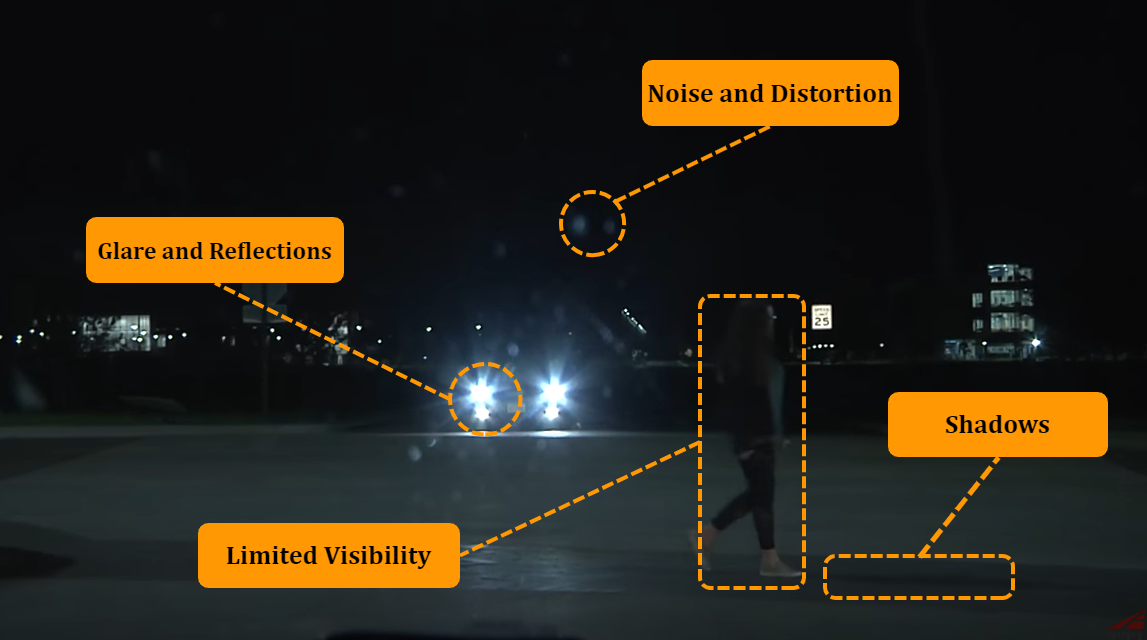}
    \caption{Challenges of detecting pedestrians at night (image taken from 2019 Traffic Safety conference nighttime visibility report by Texas A\&M Transportation Institute).}
    \label{fig_overview}
\end{figure}

Although some approaches have used \ac{LiDAR} sensors, which are accurate remote sensing technologies that use laser light for distance measurement and obstacle avoidance, such sensors do not provide rich information from the surroundings \cite{wang2022lidar}.
There are many recently introduced pedestrian detection approaches such as \cite{ghari2022robust, jiang2019pedestrian, li2019deep, barba2020deep, hung2020faster, ahmed2019enhanced} that cover the task in a wide range of scenarios.
However, few recent works focus on detecting pedestrians at night and in low visibility conditions.
In low-illumination scenarios, it is much more difficult for autonomous vehicles equipped only with vision sensors to detect moving objects on the road and prevent incidents.
Fig~\ref{fig_overview} depicts how challenging the pedestrian detection task is compared to the same ordinary task during the daytime.
The mentioned fact has resulted in an increase in demand for developing computer vision algorithms that can work under various illumination conditions.

Accordingly, this survey gives a deep review of 118 state-of-the-art low-light condition pedestrian detection algorithms.
The research questions that the present work has aimed to answer are:

\begin{itemize}
  \item \textbf{RQ1} Which datasets and baselines are mainly employed in low-light pedestrian detection tasks?
  \item \textbf{RQ2} What are the current deep learning-based algorithmic trends in low-illumination pedestrian detection?
  \item \textbf{RQ3} Regarding the state-of-the-art solutions, what are the currently existing and unresolved challenges in practical large-scale applications, such as fully autonomous vehicles?
\end{itemize}

To answer these questions, the paper in hand contributes to the body of knowledge in the field by providing contributions listed below:

\begin{itemize}
  \item A survey of more than a hundred papers in the field of nighttime pedestrian detection,
  \item Review and classification of well-known baselines and datasets used for this purpose,
  \item Categorization of the state-of-the-art approaches in nighttime pedestrian detection regarding their architectural variations,
  \item Identification of the current trends and future methodologies in the field,
\end{itemize}

The rest of the paper is organized as follows:
Section~\ref{sec_surveys} reviews the currently available surveys in pedestrian detection in low-light conditions.
Section~\ref{sec_baselines} introduces the existing baselines and datasets employed by available approaches (\textbf{RQ1}).
In Section~\ref{sec_sota}, a set of recently introduced night-time pedestrian detection approaches are introduced.
Some discussions on the revealed trends and future insights in the field are presented in Section~\ref{sec_discussion} (\textbf{RQ2}, \textbf{RQ3}).
Finally, the paper concludes in Section~\ref{sec_conclude}.
\section{Related Surveys}
\label{sec_surveys}

Given the significant importance of the pedestrian detection chore in cutting-edge domains such as autonomous vehicles, an extensive collection of surveys has been publicly available.
These surveys delve into various aspects, including methodological approaches, target environmental contexts, evaluation procedures, and pre-defined presumptions.
This section offers a concise study of these existing reviews and identifies the unexplored factors within them.
This identification of gaps underscores the unique contribution that the manuscript in hand aims to provide in this field.

Chen \etal \cite{chen2021deep} analyzed various object detection methodologies along with robust feature extractors employed in the fields of vehicle and pedestrian detection.
They also employed extensive experiments on the \textit{KITTI} vision benchmark \cite{Geiger2012CVPR} as a well-known street dataset to assess the performance of the studied algorithms in terms of accuracy, inference time, memory consumption, model size, and the number of \acf{FLOPS}.
It is important to note that their research primarily focuses on the algorithmic perspective within practical frameworks, addressing the algorithms' efficiency across diverse scenarios.
Hou \etal \cite{hou2018multispectral} studied pixel-level image fusion strategies for vision-based pedestrian detection that works in all daytime/nighttime conditions and discussed efficient strategies of combining such methods with \ac{CNN}-based fusion architectures.
The primary aim of their research is to discuss a pixel-level fusion of strategies adopted from various approaches that result in better performance for multi-spectral pedestrian detection tasks.
Accordingly, their approach does not cover the future guidelines and possible strategies for such tasks.
Authors in \cite{iftikhar2022deep} studied deep learning-based methodologies employed in pedestrian detection tasks and provided informative discussions on how effective they are compared to other traditional algorithms.
Although the mentioned survey also covers night-time pedestrian detection, the comparison among various methodologies was mainly established on different datasets with low-quality and multi-spectral instances.
Other works such as \cite{chen2023occlusion, li2022occlusion} surveyed approaches that targeted occlusion and scale variance challenges in pedestrian detection.
They discussed solutions introduced in various papers that show acceptable performance in diverse conditions with occlusion, deformation, clutter, and scale difficulties.

\begin{figure}
    \centering
    \includegraphics[width=1.0\columnwidth]{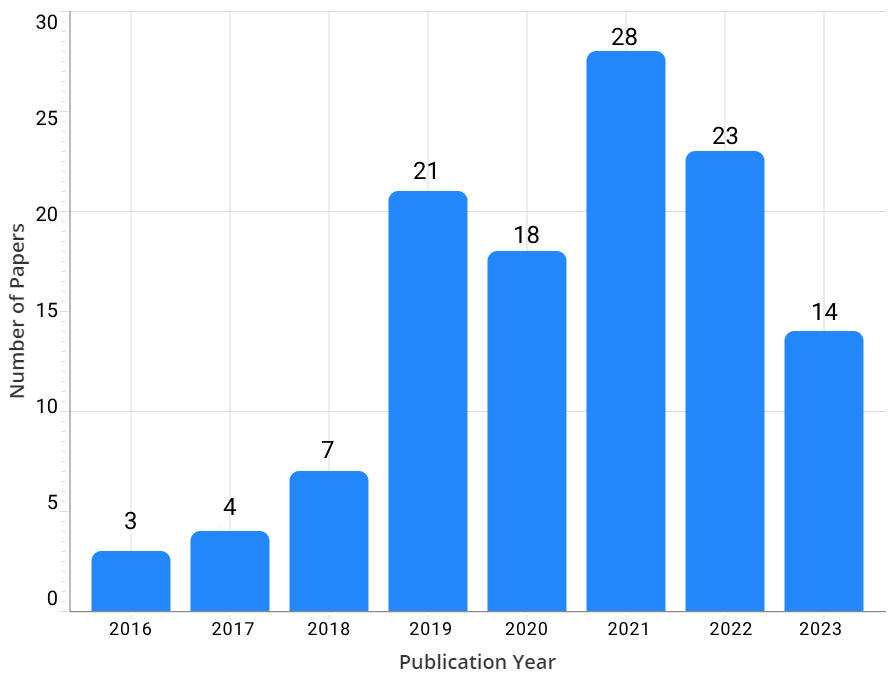}
    \caption{Distributions of the papers surveyed in the current research work that only focus on pedestrian detection at low-light scenarios from 2016-2023 (total: 118 papers).}
    \label{fig_chart_paper}
\end{figure}

Considering the introduced survey works, the present survey aims to provide specificities to set it apart from other available works, particularly regarding target scenarios and practical application use cases.
In contrast with the previous works, the survey in hand is exclusively dedicated to nocturnal pedestrian detection, shedding light on the distinctive challenges tackled under low-light conditions introduced by state-of-the-art works.
To the best of our knowledge, this work is the first study with a throughout focus on the detection of pedestrians in low-illumination (nighttime, in particular) conditions.
In order to identify pedestrian detection approaches at nighttime that produce substantial results and feature novel architectures, the authors began by gathering and screening highly read and cited works from prominent venues over recent years.
The sources included Google Scholar \footnote{\url{https://scholar.google.com/}, accessed on 30 September 2023}, as well as well-established Computer Science bibliography databases, namely Scopus \footnote{\url{https://www.dblp.org/}, accessed on 30 September 2023} and DBLP \footnote{\url{https://www.scopus.com}, accessed on 30 September 2023}.
From the publications referenced in these sources, particular attention was given to those directly related to the ones targeted for challenging low-illumination conditions and underwent further checks to ensure their alignment with the domain.
Following an in-depth exploration of the papers, they were systematically categorized based on their primary methodological solutions in addressing nighttime pedestrian detection challenges.
Fig~\ref{fig_chart_paper} depicts the distribution of papers collected, studied, and analyzed in the current survey.

\section{Benchmarking Datasets}
\label{sec_baselines}

\begin{figure*}[t]
     \centering
     \begin{subfigure}[t]{0.245\textwidth}
         \centering
         \includegraphics[width=\textwidth]{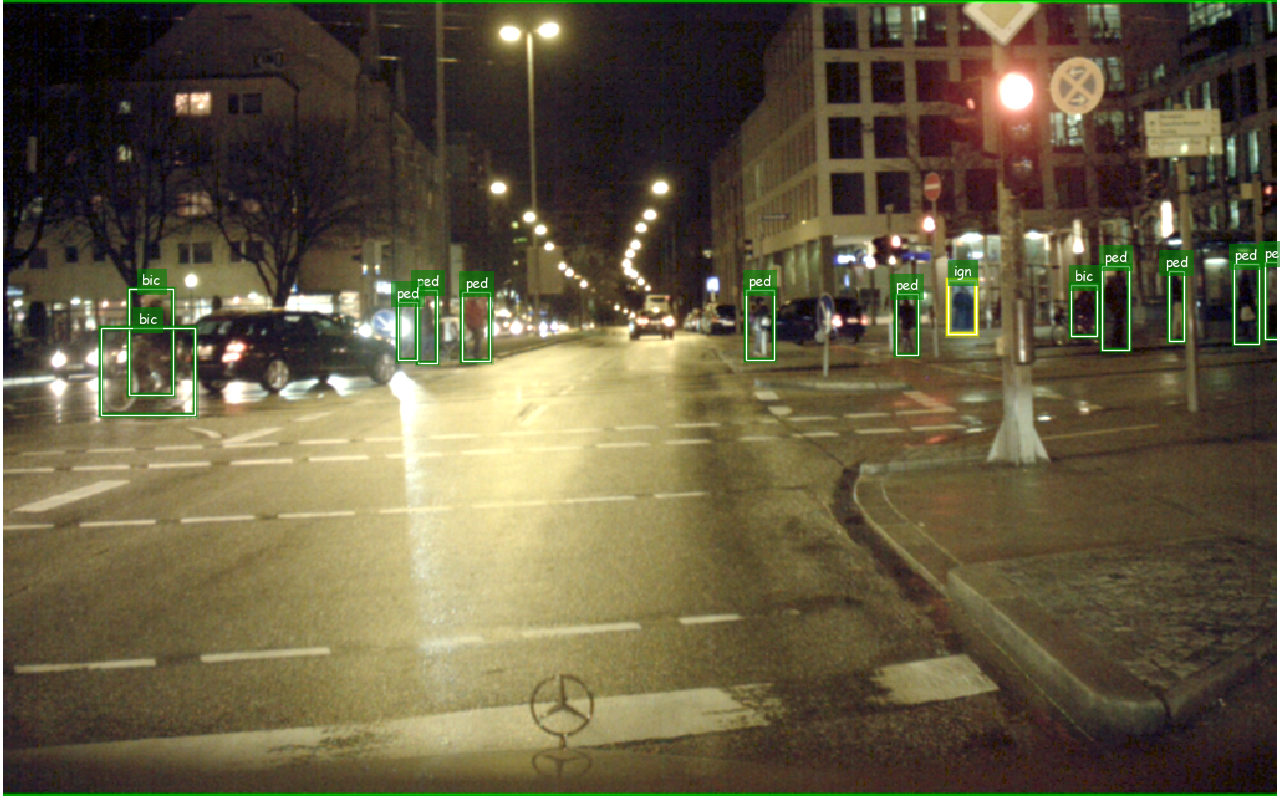}
         \caption{NightOwls \cite{neumann2019nightowls}}
         \label{fig_datasets_nightowls}
         \vspace{1em}
     \end{subfigure}
     \hfill
     \begin{subfigure}[t]{0.245\textwidth}
         \centering
         \includegraphics[width=\textwidth]{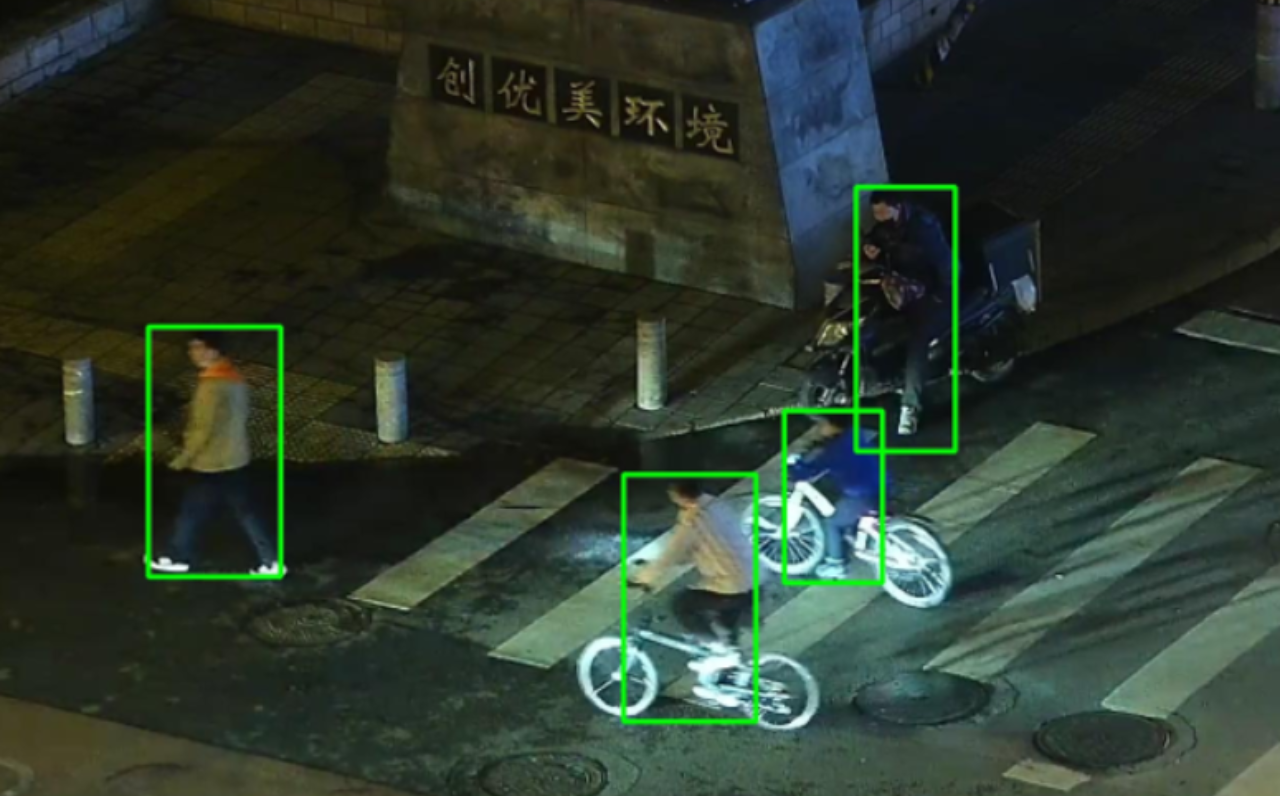}
         \caption{LLVIP \cite{jia2021llvip}}
         \label{fig_datasets_llvip}
     \end{subfigure}
     \hfill
     \begin{subfigure}[t]{0.245\textwidth}
         \centering
         \includegraphics[width=\textwidth]{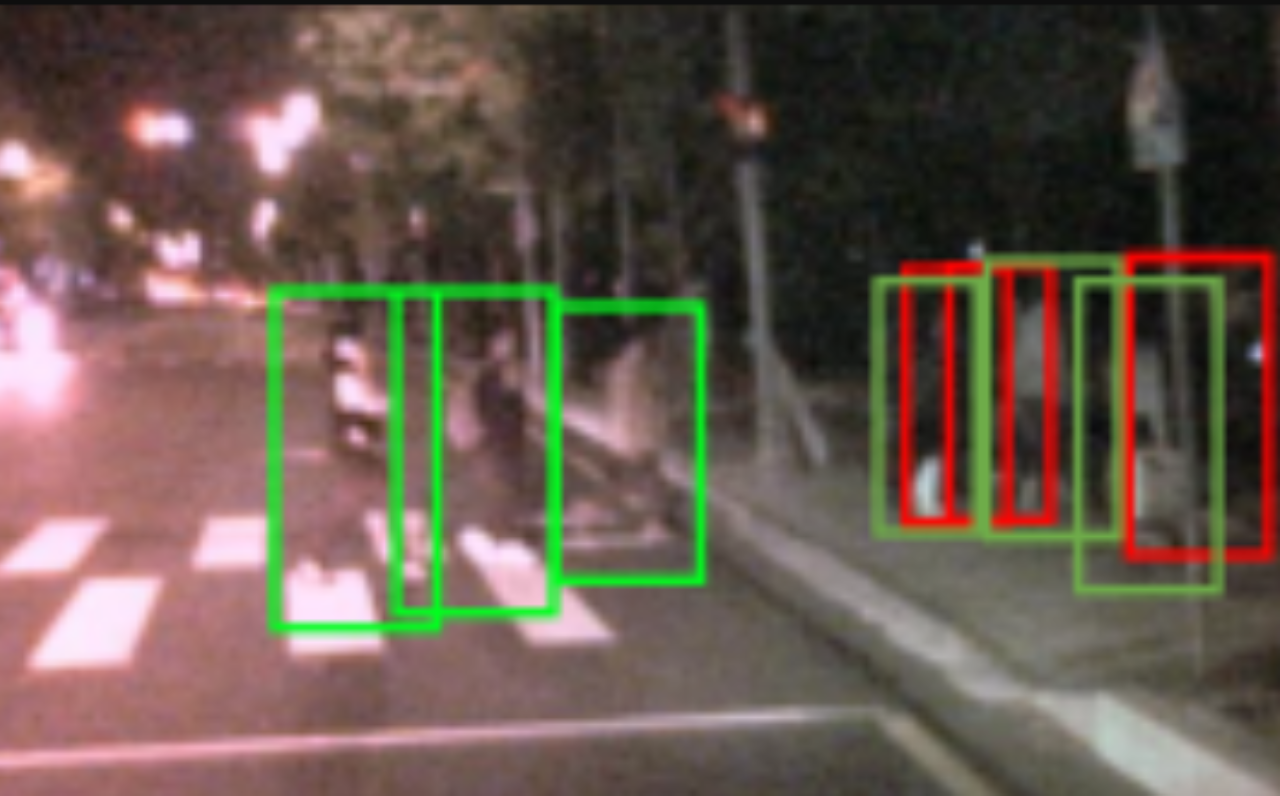}
         \caption{KAIST \cite{hwang2015multispectral}}
         \label{fig_datasets_kaist}
     \end{subfigure}
     \hfill
     \begin{subfigure}[t]{0.245\textwidth}
         \centering
         \includegraphics[width=\textwidth]{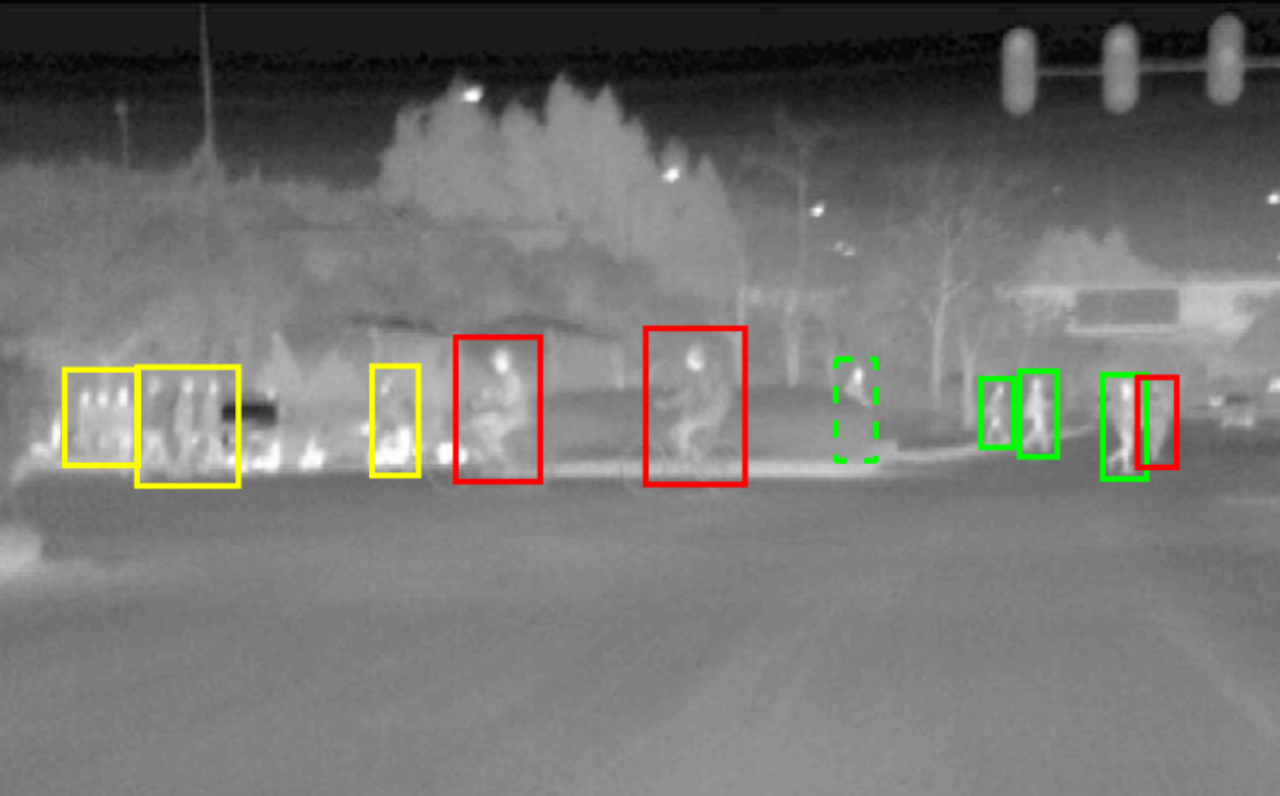}
         \caption{SCUT \cite{xu2019benchmarking}}
         \label{fig_datasets_scut}
     \end{subfigure}
     \hfill
     \begin{subfigure}[t]{0.245\textwidth}
         \centering
         \includegraphics[width=\textwidth]{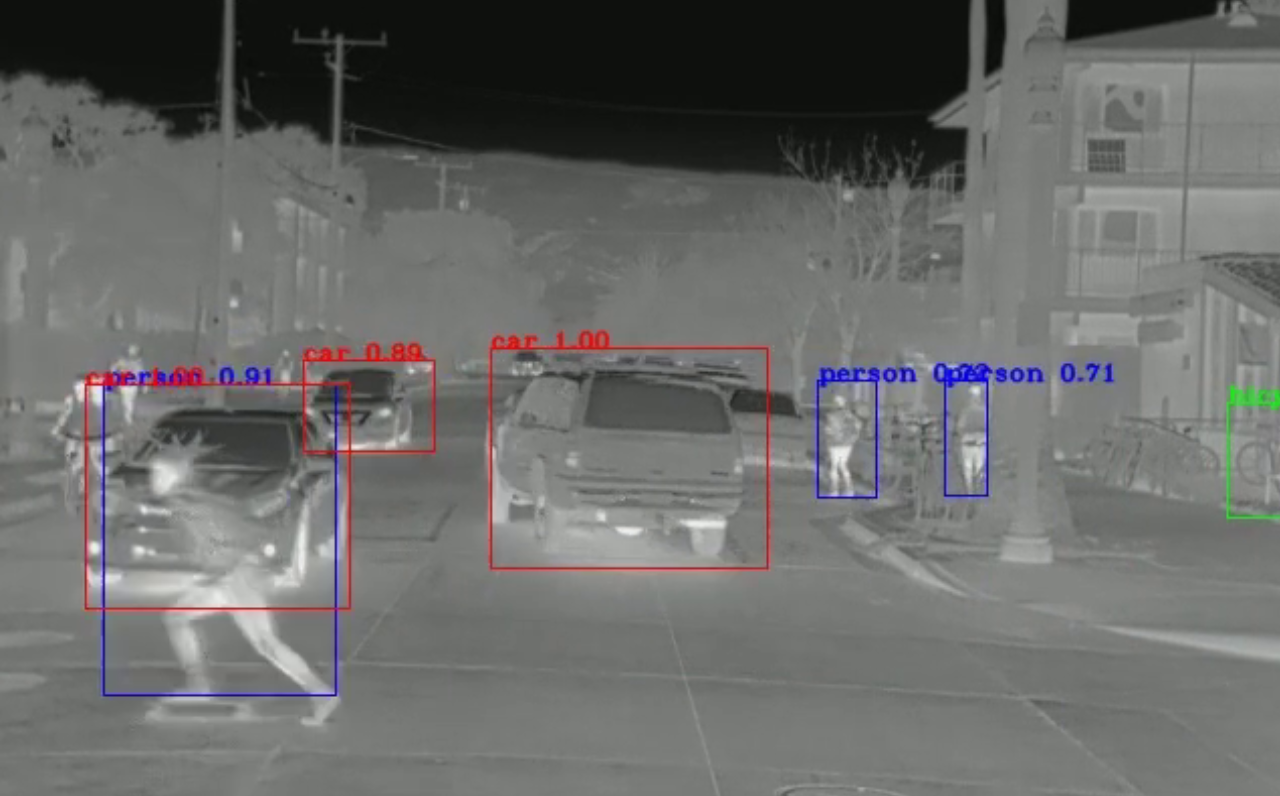}
         \caption{FLIR \cite{zhang2020multispectral}}
         \label{fig_datasets_flir}
     \end{subfigure}
     \hfill
     \begin{subfigure}[t]{0.245\textwidth}
         \centering
         \includegraphics[width=\textwidth]{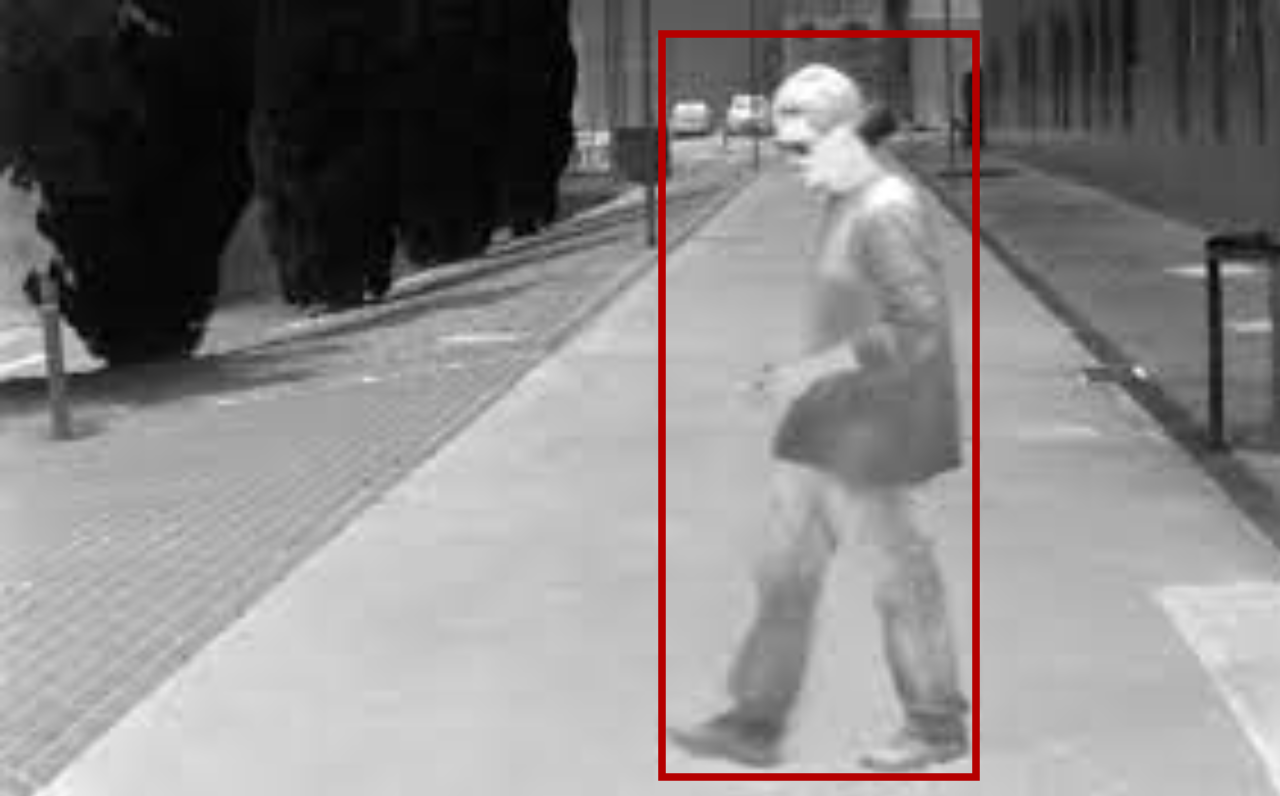}
         \caption{CVC-09 \cite{socarras2013adapting}}
         \label{fig_datasets_cvc09}
     \end{subfigure}
     \hfill
     \begin{subfigure}[t]{0.245\textwidth}
         \centering
         \includegraphics[width=\textwidth]{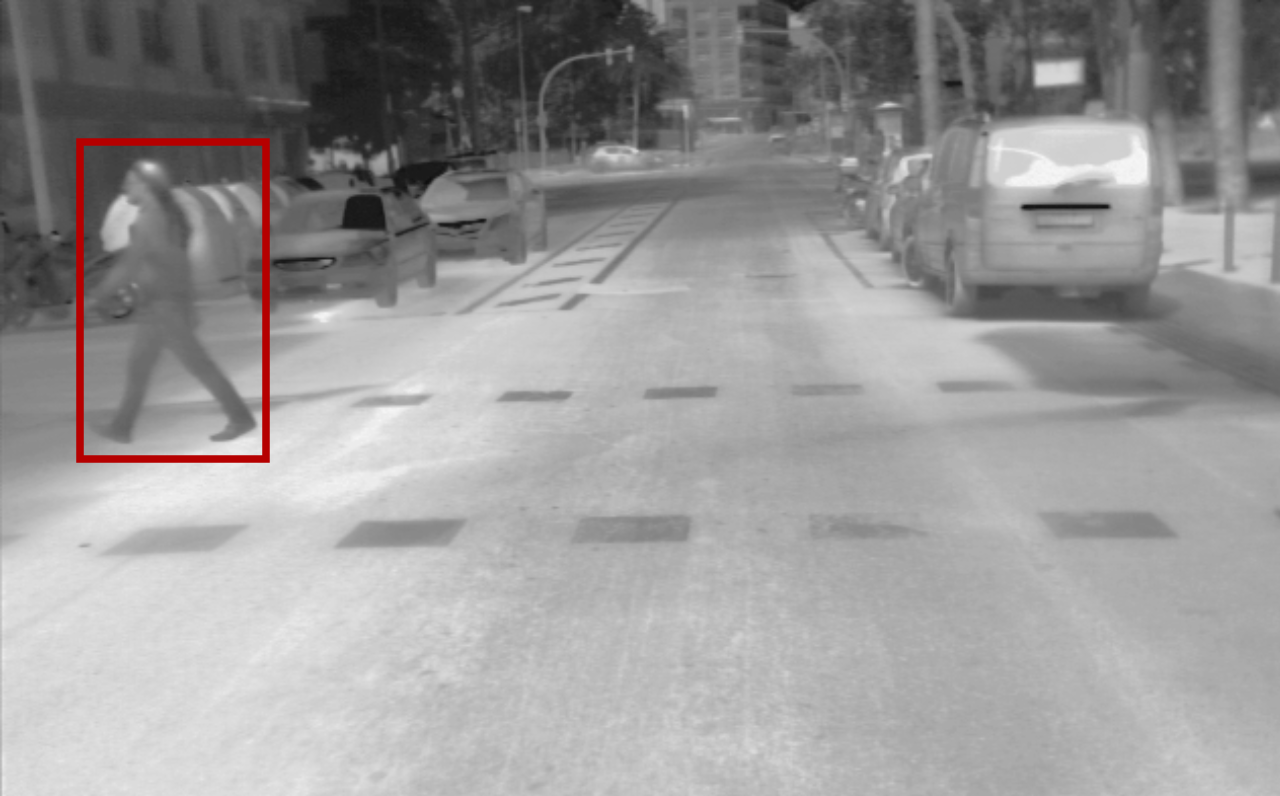}
         \caption{CVC-14 \cite{gonzalez2016pedestrian}}
         \label{fig_datasets_cvc14}
     \end{subfigure}
     \hfill
     \begin{subfigure}[t]{0.245\textwidth}
         \centering
         \includegraphics[width=\textwidth]{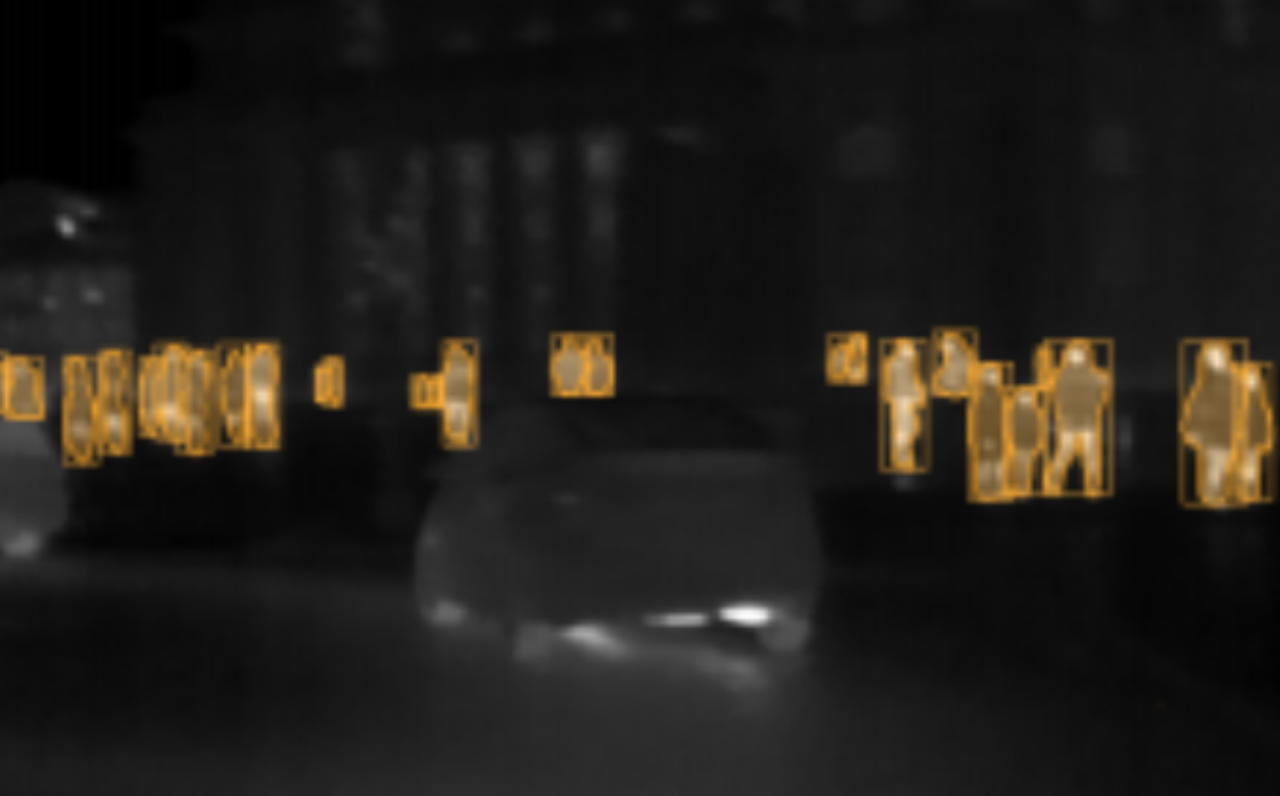}
         \caption{ZUT \cite{tumas2020pedestrian}}
         \label{fig_datasets_zut}
     \end{subfigure}
     \hfill
     \begin{subfigure}[t]{0.245\textwidth}
         \centering
         \includegraphics[width=\textwidth]{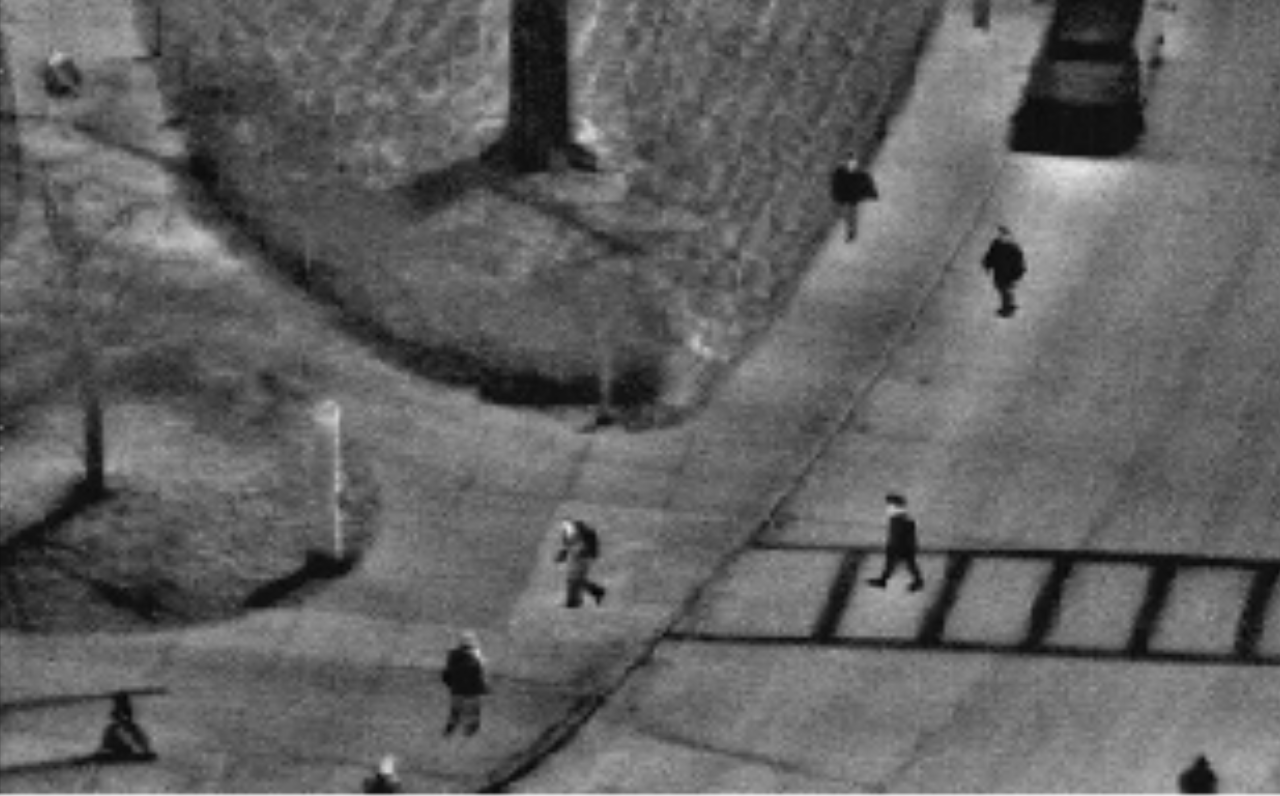}
         \caption{OSU \cite{davis2005two}}
         \label{fig_datasets_osu}
     \end{subfigure}
     \hfill
     \begin{subfigure}[t]{0.245\textwidth}
         \centering
         \includegraphics[width=\textwidth]{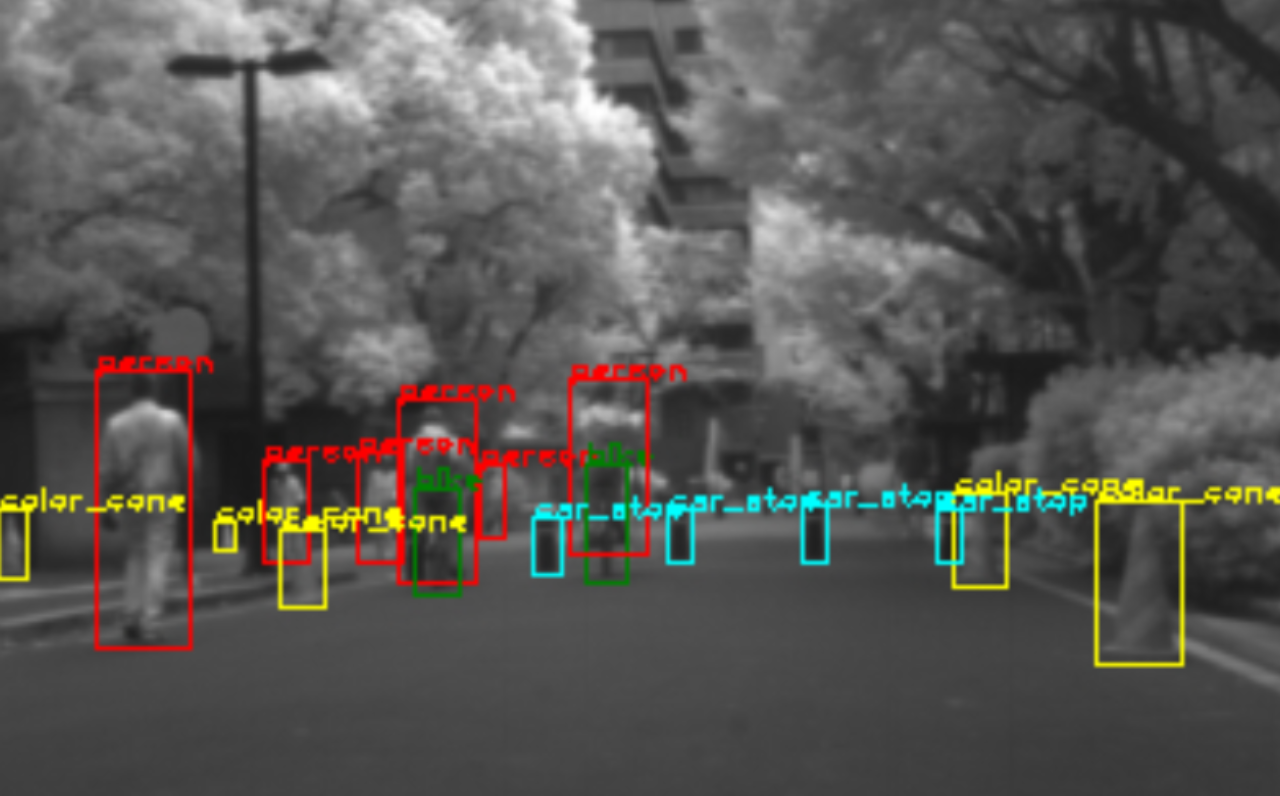}
         \caption{UTokyo \cite{takumi2017multispectral}}
         \label{fig_datasets_utokyo}
     \end{subfigure}
     \hfill
     \begin{subfigure}[t]{0.245\textwidth}
         \centering
         \includegraphics[width=\textwidth]{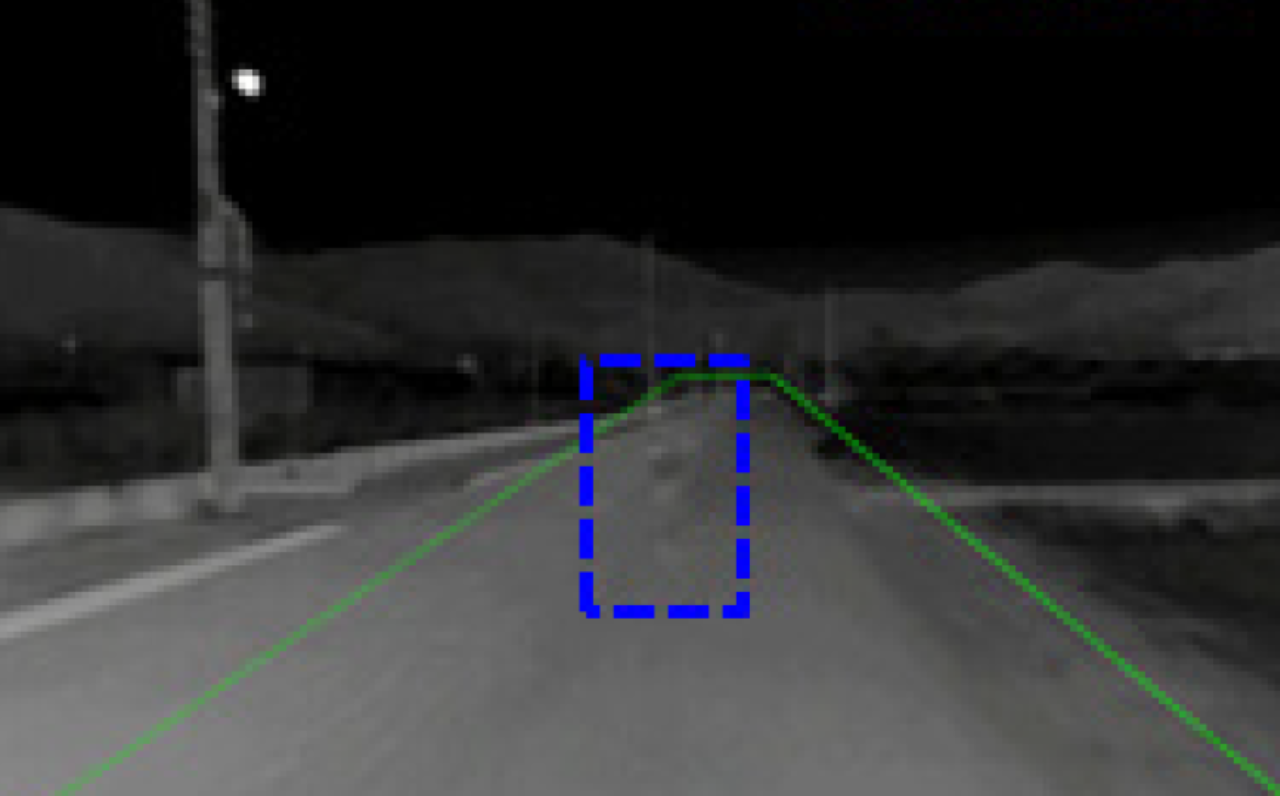}
         \caption{KMU \cite{jeong2016early}}
         \label{fig_datasets_kmu}
     \end{subfigure}
     \hfill
     \begin{subfigure}[t]{0.245\textwidth}
         \centering
         \includegraphics[width=\textwidth]{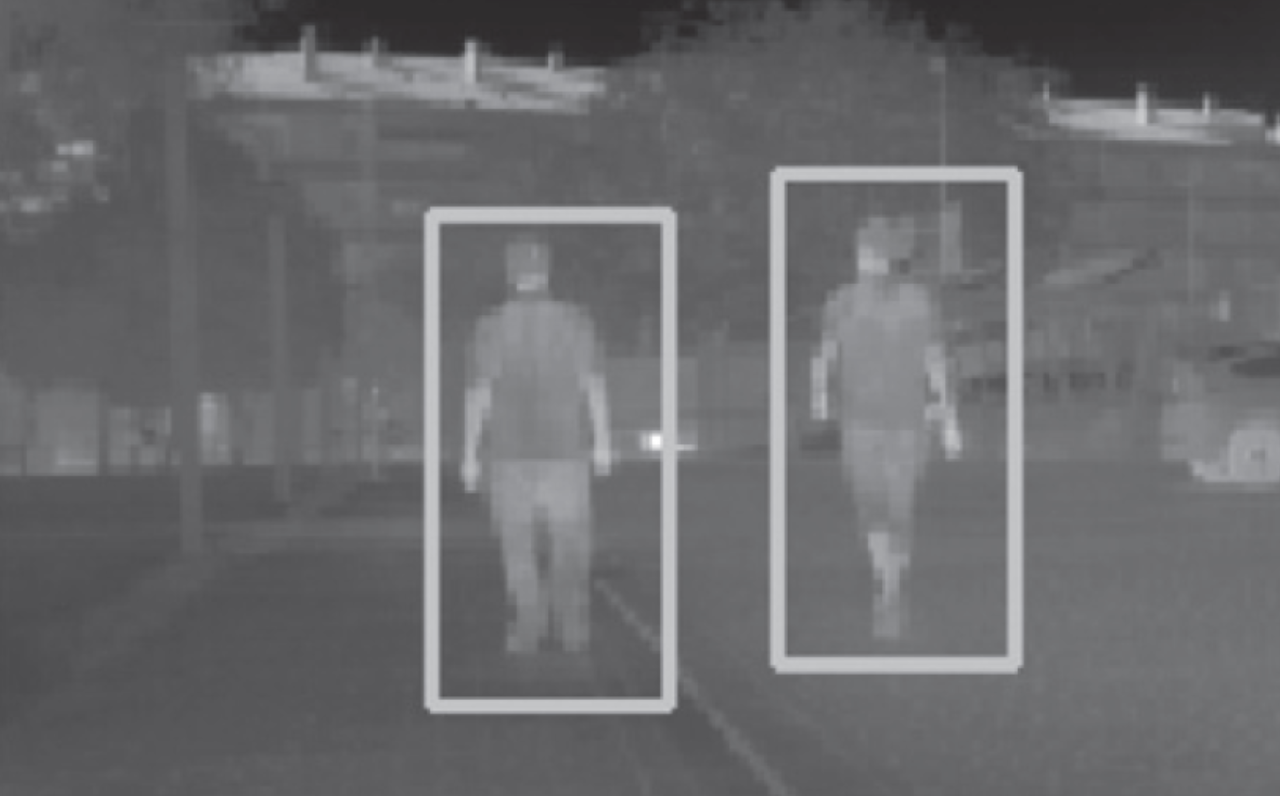}
         \caption{LSI \cite{olmeda2013pedestrian}}
         \label{fig_datasets_lsi}
     \end{subfigure}
     \hfill
     \begin{subfigure}[t]{0.245\textwidth}
         \centering
         \includegraphics[width=\textwidth]{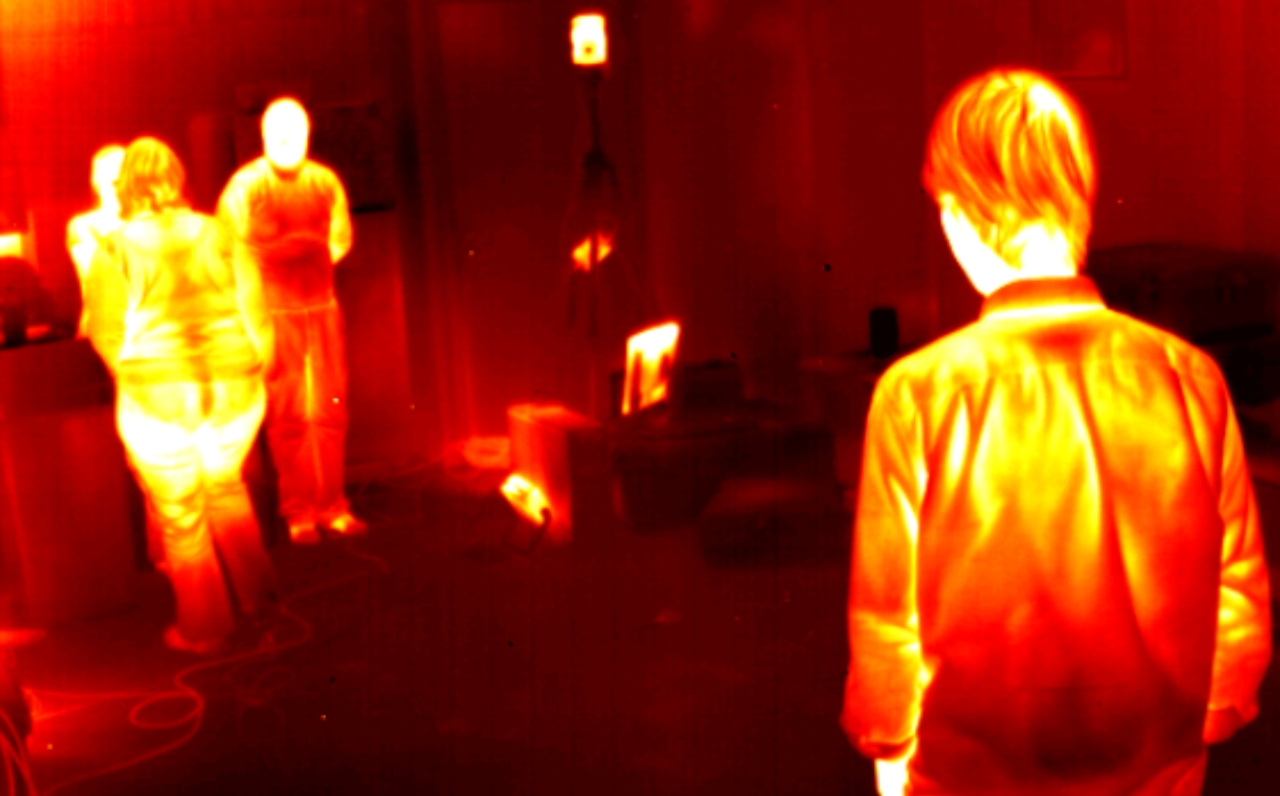}
         \caption{TIV \cite{wu2014thermal}}
         \label{fig_datasets_tiv}
     \end{subfigure}
     \hfill
     \begin{subfigure}[t]{0.245\textwidth}
         \centering
         \includegraphics[width=\textwidth]{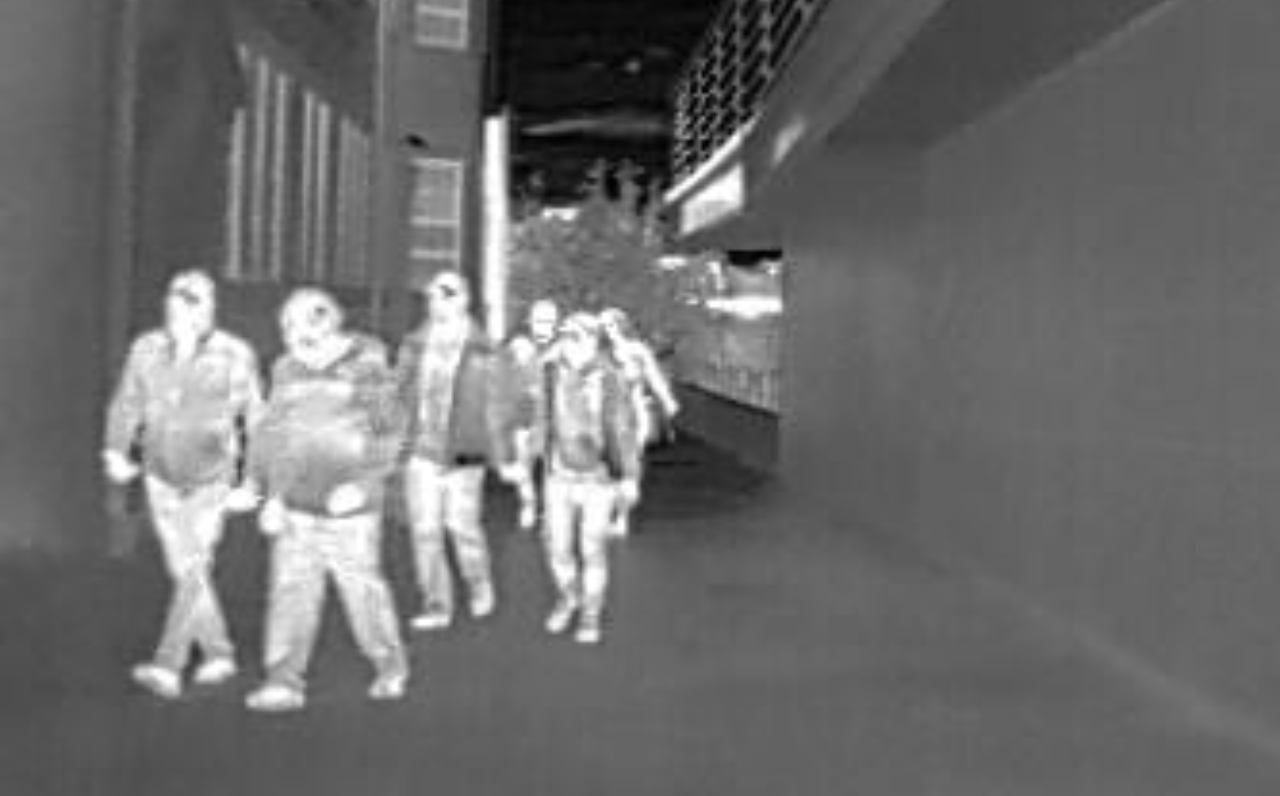}
         \caption{CAMEL \cite{gebhardt2018camel}}
         \label{fig_datasets_camel}
     \end{subfigure}
     \hfill
     \begin{subfigure}[t]{0.245\textwidth}
         \centering
         \includegraphics[width=\textwidth]{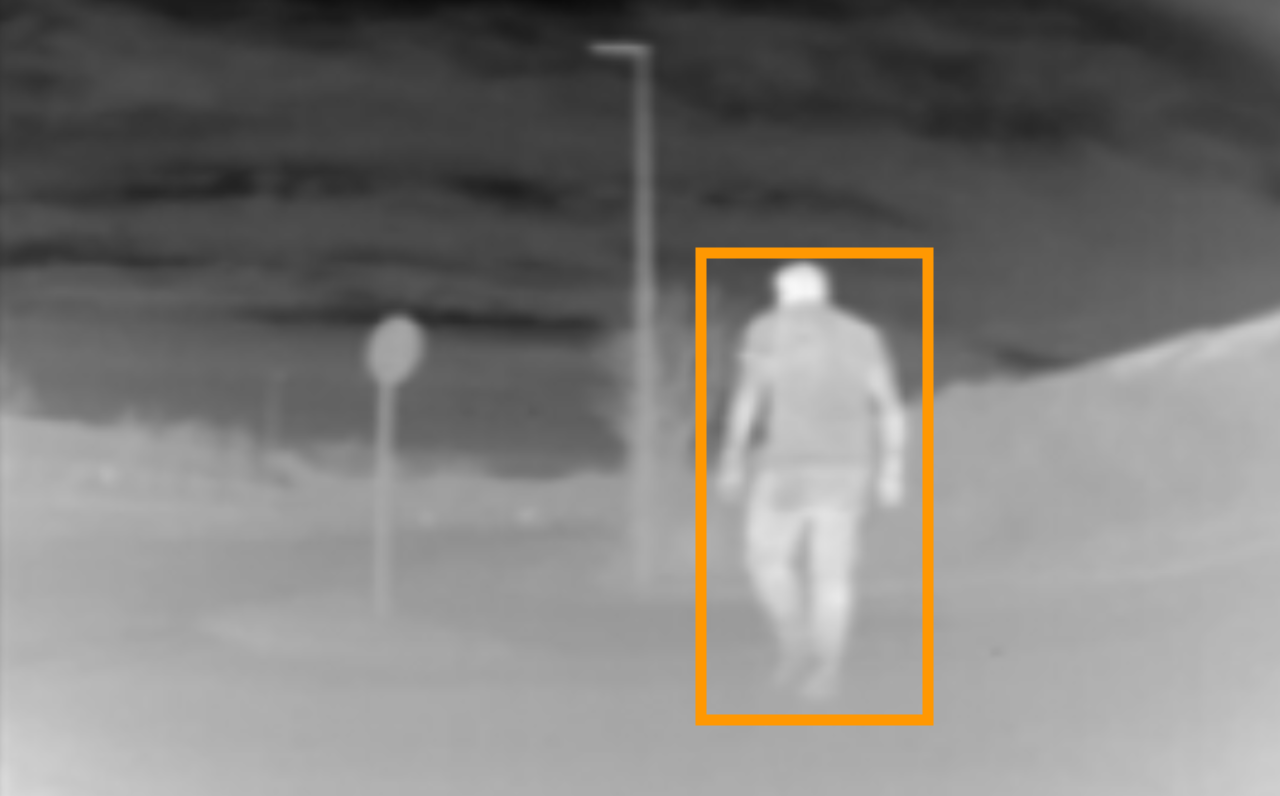}
         \caption{C3I \cite{farooq2022evaluation}}
         \label{fig_datasets_c3i}
     \end{subfigure}
     \hfill
     \begin{subfigure}[t]{0.245\textwidth}
         \centering
         \includegraphics[width=\textwidth]{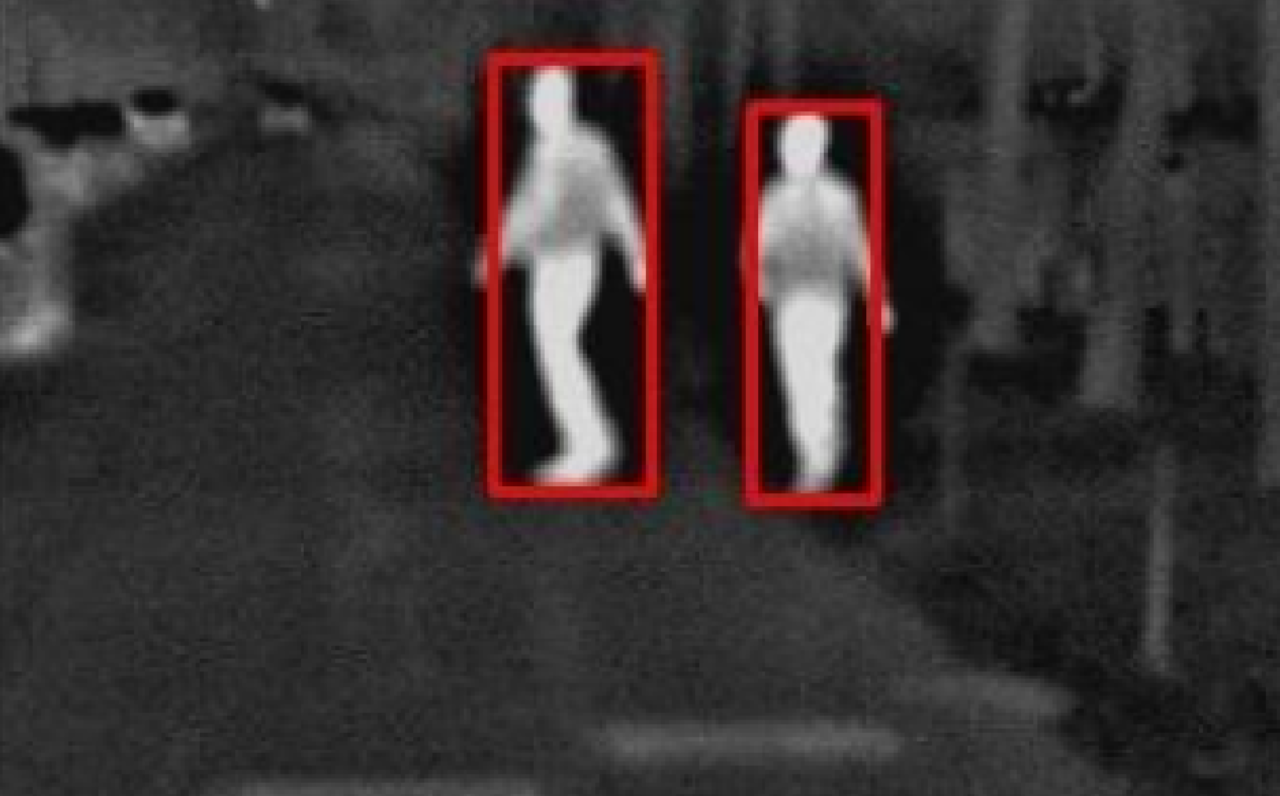}
         \caption{NTPD \cite{karol2015video}}
         \label{fig_datasets_ntpd}
     \end{subfigure}
     \caption{Instances of some datasets introduced for nighttime pedestrian detection. It should be noted that the collected image or video sequences were captured using various sensors.}
     \label{fig_datasets}
\end{figure*}

Evaluation and development of pedestrian detection algorithms highly depend on providing proper data with annotated images/videos containing pedestrian instances.
Such datasets should be well-annotated and cover diverse samples of pedestrian shots captured in real-world scenarios with various poses, occlusion levels, appearances, \etc to be considered appropriate for accurate training, testing, and validation stages.
In this regard, this section collects various standard datasets for pedestrian detection at night that can be used for training and evaluation along with facilitating benchmark creation.

\begin{table*}[t]
    \centering
    \caption{Various datasets collected for pedestrian detection at night, sorted based on their publication year. Accordingly, dataset instances are collected using various sensors in different spectral ranges, \ie \acf{NIR}, \acf{MIR}, and \acf{FIR}.}
    \begin{tabular}{l | c | cccccc | cccc }
        \toprule
            & \textbf{Metadata} & \multicolumn{6}{c|}{\textbf{Data}} & \multicolumn{4}{c}{\textbf{Sensor}} \\
            \cmidrule{2-12}
            \textbf{Dataset} & \textit{Published} & \textit{\#Videos} & \textit{\#Frames} & \textit{\#Pedestrians} & \textit{Resolution} & \textit{Frame-rate$^{\mathrm{*}}$} & \textit{Bit depth} &
            \textit{RGB} & \textit{\ac{NIR}} & \textit{\ac{MIR}} & \textit{\ac{FIR}} \\
            \midrule
                \acs{OSU} \cite{davis2005two} & 2005 & 10 & $\sim$1.9k & 984 & 360$\times$240 & 30 & 8 & & \cmark & \cmark & \cmark \\
                \acs{LITIV} \cite{torabi2012iterative} & 2012 & 9 & $\sim$6.3k & - & 320$\times$240 & 30 & 8 & \cmark & \cmark & \cmark & \cmark \\
                CVC-09 \cite{socarras2013adapting} & 2013 & 2 & $\sim$11k & $\sim$14k & 640$\times$480 & - & - & & & & \cmark\\
                \acs{LSI-FIR} \cite{olmeda2013pedestrian} & 2013 & 13 & $\sim$15.2k & $\sim$16.1k  & 164$\times$129 & - & 14 & & & & \cmark \\
                \acs{TIV} \cite{wu2014thermal} & 2014 & 16 & $\sim$63.7k & - & 512$\times$512 & 30 & 16 & & \cmark & \cmark & \cmark \\
                \acs{KAIST} \cite{hwang2015multispectral} & 2015 & 12 & $\sim$95k & $\sim$103k & 640$\times$480 & 20 & 8 & \cmark & \cmark & \cmark & \cmark \\
                \acs{NTPD} \cite{karol2015video} & 2015 & - & $\sim$22k & - & 64$\times$128 & - & - & & \cmark & \cmark & \cmark \\
                CVC-14 \cite{gonzalez2016pedestrian} & 2016 & 4 & $\sim$8.5k & $\sim$9.3k & 640$\times$512 & 10 & - & \cmark & & & \cmark \\
                \acs{KMU} \cite{jeong2016early} & 2016 & 23 & $\sim$12.9k & - & 640$\times$480 & 30 & 24 & & & & \cmark \\
                UTokyo \cite{takumi2017multispectral} & 2017 & - & $\sim$7.5k & $\sim$2k & 640$\times$480 & 1 & - & \cmark & \cmark & \cmark & \cmark \\
                CAMEL \cite{gebhardt2018camel} & 2018 & 26 & $\sim$43k & $\sim$80k & 336$\times$256 & 30 & 24 & \cmark & \cmark & \cmark & \\
                NightOwls \cite{neumann2019nightowls} & 2018 & 40 & $\sim$279k & $\sim$42k & 1024$\times$640 & 15 & - & \cmark & & & \\
                \acs{SCUT} \cite{xu2019benchmarking} & 2018 & 21 & $\sim$211k & $\sim$477k & 720$\times$576 & 25 & 8 & & & & \cmark \\
                YU FIR \cite{kim2018pedestrian1} & 2018 & - & $\sim$2.8k & $\sim$9.3k & 640$\times$480 & 30 & 14 & & & & \cmark \\
                \acs{FLIR} \cite{flir} & 2020 & - & $\sim$10.2k & $\sim$28.1k & 640$\times$512 & 24 & 16 & & & & \cmark \\
                \acs{ZUT} \cite{tumas2020pedestrian} & 2020 & - & $\sim$110k & $\sim$80k & 640$\times$480 & 30 & 16 & & \cmark & \cmark & \cmark \\
                \acs{LLVIP} \cite{jia2021llvip} & 2021 & 26 & $\sim$33.6k & - & 1080$\times$720 & 1 & 24 & \cmark & \cmark & \cmark & \cmark \\
                C3I \cite{farooq2022evaluation} & 2022 & 6 & $\sim$39k & - & 640$\times$480 & 30 & 8 & & \cmark & \cmark & \cmark \\
        \bottomrule
        \multicolumn{3}{l}{$^{\mathrm{*}}$presented in \textit{frames per second (fps)}}
    \end{tabular}
    \label{tbl_dataset}
\end{table*}

\begin{figure*}
    \centering
    \includegraphics[width=0.85\textwidth]{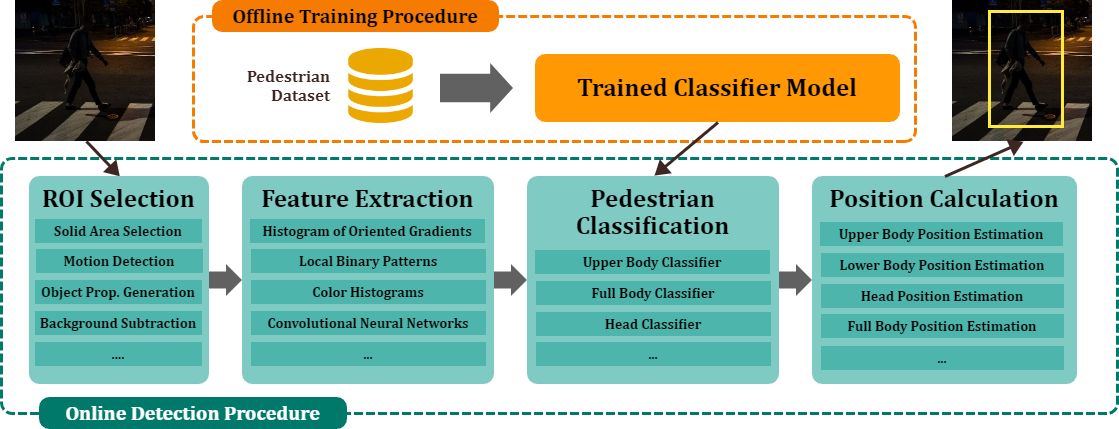}
    \caption{The overall diagram of night-time pedestrian detection methodologies.}
    \label{fig_diagram}
\end{figure*}

\noindent\textbf{\ac{OSU} Dataset} \footnote{\url{https://vcipl-okstate.org/pbvs/bench/Data/01/download.html}} \cite{davis2005two}:
As one of the first pedestrian datasets, this thermal database provides a total number of ~$1.9$k thermal frames with a resolution of \(360\times240\), which were captured on campus and street. The \textit{OSU} comprises three different classes of objects, including persons, cars, and poles. A total of $984$ people were annotated in this dataset.  

\noindent\textbf{\ac{LITIV}} \footnote{\url{https://www.polymtl.ca/litiv/en/codes-and-datasets}} \cite{torabi2012iterative}:
The dataset has nine video sequences, each containing people in an indoor hall with various zoom settings.
The main challenges in these video sequences are the strong occlusions of objects and cluttered backgrounds.

\noindent\textbf{CVC-09 Dataset} \footnote{\url{http://adas.cvc.uab.es/elektra/enigma-portfolio/item-1/}} \cite{socarras2013adapting}:
As another well-known dataset, CVC-09 is acquired during the day and night with ~$11$k frames.
The dataset contains training and testing sets, where the day and night sequences contain $5,990$ and $5,081$ frames, respectively.

\noindent\textbf{\ac{LSI-FIR} Dataset} \footnote{\url{https://www.kaggle.com/datasets/muhammeddalkran/lsi-far-infrared-pedestrian-dataset/code}} \cite{olmeda2013pedestrian}:
This dataset is composed of classification and detection portions and contains grayscale images collected in different temperatures with varying illumination.
The classification part has $16,152$ positive samples (\ie pedestrian) and $65,440$ negative samples (\ie background), while the detection part includes $15,224$ images, categorized into $6,159$ train and $9,065$ test instances.

\noindent\textbf{\ac{TIV}} \footnote{\url{http://csr.bu.edu/BU-TIV/}} \cite{wu2014thermal}:
The dataset contains video sequences with $63,782$ annotated frames for visual processing tasks, such as detection, counting, group motion estimation, and single-view and multiple-view tracking.
Three out of sixteen sequences are mainly used for pedestrian detection, while other classes like car, runner, bicycle, and motorcycle are marked in this dataset.

\noindent\textbf{\ac{KAIST}} \footnote{\url{https://github.com/SoonminHwang/rgbt-ped-detection}} \cite{hwang2015multispectral}:
It is one of the first multi-spectral pedestrian datasets with $95$k aligned color-thermal image pairs and $103$k dense annotation of samples.
Data was captured from various traffic scenarios in the daytime and nighttime for autonomous driving applications.
Annotations were manually added, resulting in three primary categories (person, people, and cyclist), and three occlusion levels (no-occlusion, partial-occlusion, and heavy occlusion).

\noindent\textbf{\ac{NTPD} Dataset} \cite{karol2015video}:
It contains a set of pedestrian images recorded by an active night vision system.
The dataset contains $1,998$ positive and $8,730$ negative in the training set and $2,370$ positive and $9,000$ negative samples in the testing set.

\noindent\textbf{CVC-14 Dataset} \footnote{\url{http://adas.cvc.uab.es/elektra/enigma-portfolio/cvc-14-visible-fir-day-night-pedestrian-sequence-dataset/}} \cite{gonzalez2016pedestrian}:
An extended version of the CVC-09 dataset, titled CVC-14, was introduced later to facilitate the challenges of automated driving.
It contains video sequences of grayscale visible and thermal pairs corresponding to daytime and nighttime, where the daytime and the nighttime shares are $4,401$ and $4,117$ instances, respectively.

\noindent\textbf{\ac{KMU} Dataset} \footnote{\url{https://cvpr.kmu.ac.kr/KMU-SPC.html}} \cite{jeong2016early}:
As a dataset captured using a \ac{FIR} camera mounted on a vehicle driving in the summer nights for pedestrian detection, it contains three types of videos regarding the driving speed ($20$-$30$ km/h).
It also covers pedestrians with different activities and poses, such as walking, running, and crossing the road.
\ac{KMU} has $4,474$ positive and $3,405$ negative frames in the training set and $5,045$ frames in the testing set.

\noindent\textbf{UTokyo} \footnote{\url{http://www.mi.t.u-tokyo.ac.jp/projects/mil_multispectral/}} \cite{takumi2017multispectral}:
This multi-spectral dataset contains RGB, \ac{NIR}, \ac{MIR}, and \ac{FIR} images collected in a university for object detection in automated driving, including person, car, and bike.
It contains $7,512$ images, where $3,740$ was taken during the daytime and the rest at nighttime.

\noindent\textbf{CAMEL} \footnote{\url{https://camel.ece.gatech.edu/}} \cite{gebhardt2018camel}:
The dataset provides visible-infrared video sequences for multiple object detection and tracking, where $43$k visible-infrared image pairs are annotated with four different object classes, including person, bike, vehicle, and motorcycle.
CAMEL covers various real-world scenes, occluded targets, and different illumination conditions.

\noindent\textbf{NightOwls} \footnote{\url{https://www.nightowls-dataset.org/}} \cite{neumann2019nightowls}:
This dataset targets the research on pedestrian detection at night and contains videos recorded in seven cities across Germany, the Netherlands, and the United Kingdom.
It contains $279k$ frames with $42k$ pedestrians that have been manually labeled.
Three primary labels (\ie far, medium, and near) have been assigned to the pedestrians to categorize them based on the distance they had from the vehicle during data acquisition.
Additionally, frame brightness levels (low, medium, and high) and pedestrian pose (frontal and sideways) are other classification metrics employed in NightOwls.

\noindent\textbf{\acf{SCUT} Dataset} \footnote{\url{https://github.com/SCUT-CV/SCUT_FIR_Pedestrian_Dataset}} \cite{xu2019benchmarking}:
A large-scale nighttime pedestrian dataset proposed by Xu \etal to motivate more attempts toward the task of on-road \ac{FIR} pedestrian detection.
The dataset contains approximately $11$ hours-long image sequences with $211$k annotated frames and a total of $477$k bounding boxes for $7$k unique pedestrians.
\textit{SCUT} groups pedestrians into three subsets, including near-scale (\ie $\sim80$ pixels), medium-scale (\ie $\sim30$ to $\sim80$ pixels), and far-scale (\ie less than $30$ pixels) subset based on the range of imaging distances.

\noindent\textbf{YU FIR} \cite{kim2018pedestrian1}:
This seasonal temperature-based pedestrian detection dataset is captured on campus and urban traffic roads.
The temperature was calibrated from -40°C to 150°C and used as the thermal infrared data for pedestrian detection.
YU FIR contains a total of $2,802$ frames with $1,803$ and $575$ positive images in the training set and test set, respectively.

\noindent\textbf{\ac{FLIR} Dataset} \footnote{\url{https://www.flir.com/oem/adas/adas-dataset-form/}} \cite{flir}:
This multi-spectral dataset was collected for \acf{ADAS} during daytime and nighttime.
It contains visible-thermal image pairs, some of which are not aligned, and the rest contain $5$k multi-spectral pairs for training and testing.
This version contains three frequent object categories, including persons, bicycles, and cars.

\noindent\textbf{\ac{ZUT} Dataset} \footnote{\url{https://ieee-dataport.org/open-access/zut-fir-adas}} \cite{tumas2020pedestrian}:
It is a thermal dataset recorded in four European countries during diverse weather conditions, including sunny, foggy, heavy rain, light rain, and cloudy.
The dataset contains $110$k frames with $80$k pedestrian annotations and provides synchronized \ac{CAN bus} data, including brake pedal status, driving speed, and outside temperature for \ac{ADAS}.

\noindent\textbf{\ac{LLVIP} Dataset} \footnote{\url{https://github.com/bupt-ai-cz/LLVIP/}} \cite{jia2021llvip}:
The dataset is recorded by a binocular camera containing visible light and \ac{IR} sensors.
Targeting low-illumination surveillance tasks, the dataset contains $15k$ pairs of visible-infrared images.
The annotations of \ac{IR} and visible-light images are the same due to the similar resolution and \ac{FoV} of the cameras.

\noindent\textbf{C3I Thermal Automotive Dataset} \footnote{\url{https://ieee-dataport.org/documents/c3i-thermal-automotive-dataset/}} \cite{farooq2022evaluation}:
The dataset was acquired in various environmental (\ie roadside, industrial town, alley, and downtown) and weather (\ie cloudy, foggy, windy, and sunny weather) conditions during daytime, evening, and nighttime.
It comprises video sets with $39,770$ frames, of which $17,740$ frames are recorded in daytime, $12,640$ in evening time, and $9,390$ frames at nighttime.
The frames are annotated in six object classes: person, car, bike, bicycle, bus, and pole.

To provide a comprehensive introduction to the datasets at hand for nighttime pedestrian detection, Fig.~\ref{fig_datasets} depicts some of their instances.
Additionally, Table~\ref{tbl_dataset} provides an in-depth description, facilitating a deeper understanding of their attributes and characteristics.
It should be noted that these datasets have been curated in a way that encompasses a wide range of scenarios, including various lighting conditions, diverse pedestrian poses, occlusions, and complicated backgrounds.
Such diversities ensure that provided data can serve as valuable resources for evaluating algorithms under real-world conditions.
\section{State of the Art}
\label{sec_sota}

\begin{figure*}
    \centering
    \includegraphics[width=0.85\textwidth]{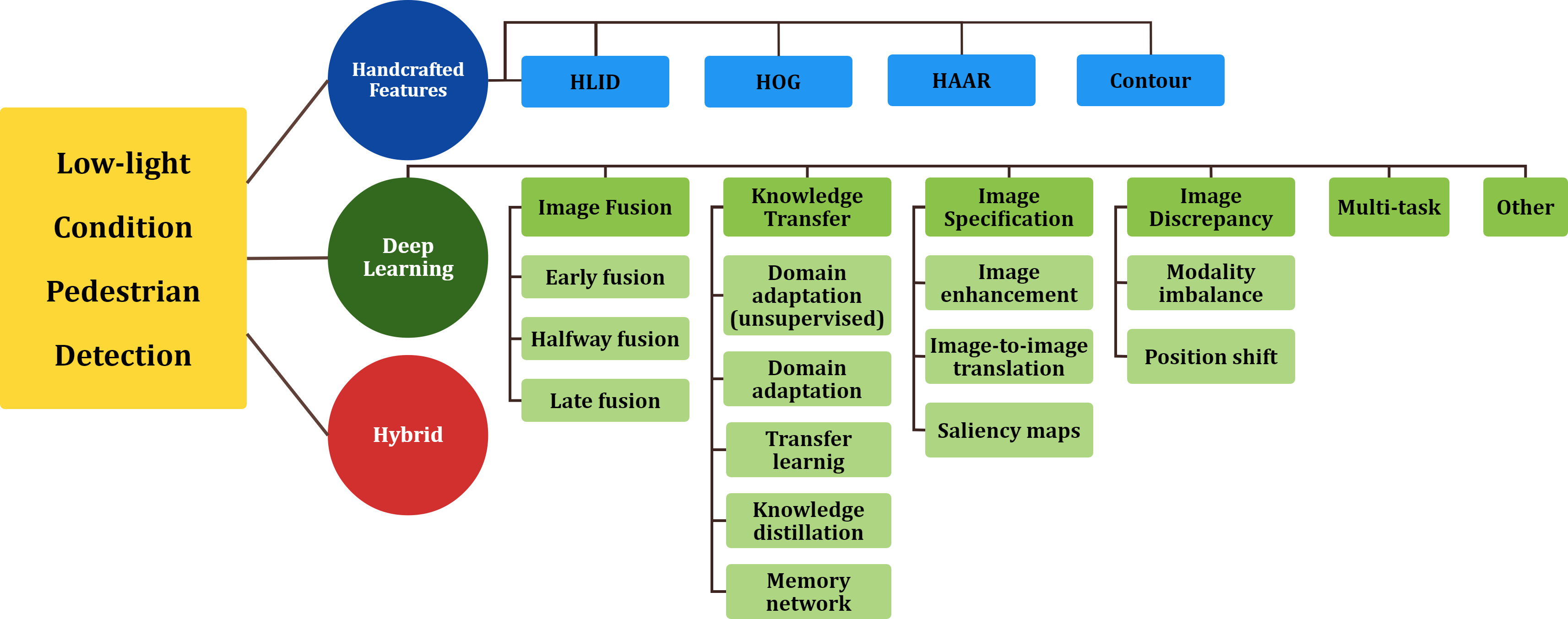}
    \caption{The primary classification of different nighttime pedestrian detection methodologies considered in this survey.}
    \label{fig_tree}
\end{figure*}

When considering the pedestrian detection methodologies for nighttime and low-illumination conditions, one of the primary architectures that come to mind for designing such frameworks is to include an \textit{offline training} procedure that utilizes a dedicated pedestrian images dataset to train a classification model.
In this regard, the model learns hidden patterns and characteristics specific to pedestrians in darker environments, enabling it to distinguish them from other objects or background elements.
Although the mentioned architecture can be very beneficial, it should be noted that learning-based methodologies are not always used for this task.
Fig.~\ref{fig_diagram} shows these typical stages for pedestrian detection at night and how they are connected to each other.
We can see the typical stages that form the foundation and serve as critical components in identifying pedestrians, irrespective of the specific methodology employed.
Whether the framework employs handcrafted features, machine learning models, or a mixture of both, it typically encompasses standard stages, including \ac{ROI} selection (\ie identifying potential pedestrian regions within an image), visual feature extraction (\ie capturing relevant information from the selected \acp{ROI}), pedestrian classification (\ie using the features to classify the detected regions as pedestrians or non-pedestrians), and position calculation (\ie determining the precise location of the detected pedestrians).

In this survey, nighttime pedestrian detection approaches based on the underlying techniques and methodologies employed have been categorized into three distinct groups, including handcrafted features, deep learning, and hybrid methods.
To understand the diverse strategies, along with their set of advantages and limitations, this section provides an in-depth study of the state-of-the-art works classifiable in the mentioned three categories.

\subsection{Handcrafted Features Approaches}
\label{subsec_handcraft}

Handcrafted features contain manual design and selection of particular visual features from the input image/frame.
Extracting information using such features has the advantage of simplicity, transparency, explainability, and the ability to provide consistent results across similar scenarios.
They generally require lower computational costs and can still work well when there is no access to annotated data.
However, they can have adaptability problems regarding their domain expertise and may reach a \textit{performance ceiling}, making them difficult to be improved over a certain point.
Considering these trade-offs, handcrafted features can still be found in many pedestrian detection frameworks under challenging illumination scenarios.

Regarding the intrinsic characteristics of handcrafted features, the majority of approaches in this category require employing thermal data.
As one of the first works in this category, Davis \etal \cite{davis2005two} used a combination of generalized person template derived from \acf{CSM} and background subtraction to identify pedestrians' locations in thermal frames.
Then, an \textit{AdaBoost} classifier could validate the candidate regions, which adaptively adjusts the filters from the gradient information of training instances.
While the template-based method brings about a quick screening procedure, it is considered a challenging methodology to detect groups of people in the scene.
Similarly, Nowosielski \etal \cite{nowosielski2020embedded} presented a HAAR and Adaboost-based night-vision framework to identify humans in thermal images.
The proposed algorithm processed all frames independently and without the aggregation mechanism, which increases the false positive rate due to incorrect recognition of the region as a person.
An approach titled \ac{TIR-ACF} introduced in \cite{kim2018pedestrian1} employs a thermal normalization methodology to factor in the maximum human body temperature for pedestrian detection.
However, the experimental environment of this normalization strategy only includes a specific temperature range of small distant targets.
As a more complicated methodology, Jeong \etal \cite{jeong2016early} presented an approach based on a \ac{CaRF} classifier, low-dimensional Haar-like features, and \ac{OCS-LBPs} for detecting sudden pedestrian crossing in thermal images.
As the thermal temperature of the road is similar to or slightly higher than the pedestrians during summer nights, the concentration of this approach is on pedestrian samples in the summer season which leads to high prediction accuracy.
Kim \etal \cite{kim2019pedestrian} designed a pedestrian detector using a multi-level cascade learning algorithm and \ac{HOG} features.
They used a smartphone-based thermal camera to capture human images of indoor environments to validate their work.
Additionally, the $2$D thermal image is mapped into a $3$D space through an inverse perspective transformation method \cite{kim2015detection} to estimate the distance of the pedestrian detected from the camera.

\acl{IR} images are another source of valuable information for pedestrian detection tasks at night.
Zhou \etal \cite{zhou2020pedestrian} designed a pedestrian extraction algorithm for \ac{IR} images.
They build a global model using the weighted \ac{HLID} and texture weighted \ac{HOG} algorithms to locate potential pedestrian regions.
Then, using a head template based on the HAAR-like features and incorporating it into a local model for pedestrian head search, the global and head templates are combined to identify pedestrians.
As another approach, Khalifa \etal \cite{khalifa2020pedestrian} introduced a foreground detection framework that models the background’s global motion between consecutive frames by applying the block-matching algorithms to the \ac{ROI} to compensate for the camera motion.
They use a \ac{SVM} classifier to differentiate between the image's foreground and background.
The evaluation results on the \textit{CVC-14} show that the proposed algorithm can capture the dynamic aspect between frames in a video stream.
Shahzad \etal \cite{shahzad2021smart} suggested a new procedure for pedestrian detection, tracking, and head detection in \ac{IR} systems using template matching, Kalman filter, and HAAR cascade classifiers, respectively.
The authors confirmed that the template matching method performs better than the contour-based method for pedestrian detection, and pedestrian tracking using the Kalman filter has the highest error rate.
Likewise, and based on visual saliency in \ac{IR} images, Cai \etal \cite{cai2017saliency} proposed a model to focus on \ac{ROI} generation along with a \ac{HLID} feature and an \ac{SVM} classifier to make a final detection.
Considering that the visual saliency-based method includes small processing regions for candidate verification, the proposed algorithm demonstrates a fast execution time.

To conclude, the approaches with handcrafted features can provide acceptable results in many cases.
However, they generally suffer from their incapability to handle complex scenarios due to their low discriminative nature and seem to act less flexibly while adapting to new scenarios.
\subsection{Deep learning approaches}
\label{subsec_dl}

Solutions based on deep learning leverage the potential of neural networks to learn and extract features from raw image data automatically.
In this regard, adaptability, versatility, generalization \wrt diverse scenarios, and high-performance results are among the expected outcomes of employing \acp{DNN}.
They also have the capability to automatically learn features and reduce the requirement for manual feature engineering, along with integrating feature extraction and detection steps to have an end-to-end learning procedure.
However, approaches in this category typically require large amounts of labeled data for training and powerful hardware due to their computationally intensive nature.
Additionally, they lack straightforward explainability, leading to challenging interpretations of their decision-making process and, thus, fine-tuning to improve their performance.

Many recent works for low-light pedestrian detection employ \acp{DNN} as an inevitable part of their algorithms.
These methodologies have been divided into categories below in this survey regarding their use case:

\subsubsection{\textbf{Image Fusion Methodologies}}
\label{subsubsec_fusion}

Image fusion refers to extracting and fusing the most significant characteristics of raw images captured by multiple sensors to generate a single image with complementary information, a compelling description of the scene, \etc \cite{zhang2021image}.
Considering the \textit{fusion} stage, \ac{CNN} fusion architectures can be divided into three primary strategies, namely, \textit{early fusion}, \textit{halfway fusion}, and \textit{late fusion}.
Fig~\ref{fig_chart_dl_apps} depicts a brief overview of different fusion architectures used in the research works.

\begin{figure}[t]
    \centering
    \includegraphics[width=0.9\columnwidth]{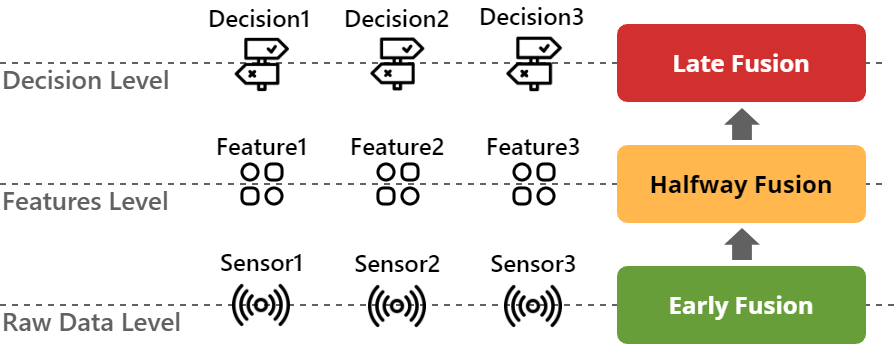}
    \caption{Various image fusion strategies used in night-time pedestrian detection approaches.}
    \label{fig_diagram_fusion}
\end{figure}

\noindent\textbf{Early fusion-based methods:}
in the context of night-time pedestrian detection, it indicates integrating visual and thermal feature maps right after the first \textit{convolutional} layer of a \ac{CNN}.
However, fusing \ac{IR} and RGB images to generate a four-channel input before feeding the network is another approach that can execute low-level feature fusion at an early stage.
As a recent work introduced in \cite{nataprawira2021pedestrian}, a \ac{YOLO}~v.3-based \cite{redmon2018yolov3} multi-spectral pedestrian detector is introduced that can use RGB, thermal, and multi-spectral images.
The mentioned approach merges the three channels of RGB and the single channel of thermal images to prepare a $4$-channels input.
The authors also evaluated a \ac{YOLO}-4L \cite{nataprawira2021pedestrian1} version to improve the detection accuracy of small-scale pedestrians on the scene.
Evaluation results on various datasets demonstrated that the methodology outperforms other image types under all lighting conditions.

\noindent\textbf{Halfway fusion-based methods:}
as a widely explored fusion strategy in recent years, the fusion operation of input modalities happens at the middle stages of a network, after the fourth convolutional layer.
As one of the halfway fusion works, Yang \etal \cite{yang2023cascaded} designed a \ac{CIEM} and a \ac{CAFFM} to enrich the pedestrian information and suppress the interference caused by background noise in the color and thermal modalities.
While \ac{CIEM} uses a \textit{spatial attention mechanism} to weigh the features combined by the cascaded feature fusion block, \ac{CAFFM} employs \textit{complementary features} to construct global features.
In \cite{zhang2020attention}, \acf{CAM} and \acf{SAM} were integrated into a multi-layer fusion \ac{CNN}, aiming to re-weigh cross-spectral features at channel-dimension and pixel-level, respectively.
Although the \ac{SAM} methodology results in reduced detection speed, the performance of the approach is substantially improved.
Zhang \etal \cite{zhang2020multispectral} suggested a new halfway fusion strategy that applies cyclical fusion and refinement operations to achieve the consistency and complementary balance of multi-spectral features by controlling the number of loops.
Based on the fact that the fused features are more discriminative than the mono-spectral ones, their main idea is to consecutively refine the spectral features with the fused ones and increase the overall feature quality.
Hence, according to the analysis on the \ac{KAIST} and FLIR dataset, the authors suggested that the number of loops should be tuned for any dataset.
A cross-modal framework based on \ac{YOLO}~v5 detector introduced in \cite{jiang2022attention} for multi-spectral pedestrian detection.
In their study, the information complementarity of RGB-thermal streams was acquired by a \acf{CFCM} to reduce the target loss.
They use an \acf{AFEFM} to fuse different modalities' essential features and suppress the background noise while strengthening the semantic information.
In another approach, and based on \ac{YOLO}~v5 lightweight network, Fu \etal \cite{fu2021adaptive} proposed an adaptive spatial and pixel-level feature fusion module, called \textit{ASPFF Net}, to obtain fusion weights of spatial positions and pixel dimensions in two feature maps.
The fusion weights are employed to re-calibrate the original feature maps of visible and \ac{IR} images to acquire multi-scale fusion feature layers.
The spatial and pixel attention mechanisms enable the \textit{ASPFF Net} to focus on learning useful information and suppress redundant information to achieve a fast prediction speed of $35$ \ac{fps} and lower \ac{MR} on the night subset of the \ac{KAIST} dataset.
A \ac{MLF-FRCNN} was proposed by \cite{deng2021pedestrian}, which employs \ac{FPN} and \ac{RPN} as two parallel feature extractors to deal with pedestrian samples with different scales.
As a two-stage multi-spectral pedestrian detector, the \ac{MLF-FRCNN} achieves a running time of 0.14 seconds per frame and the highest \ac{AP} in detecting various pedestrian scales.
In \cite{liu2016multispectral}, four variants of fusion models have been designed at different stages, titled low-level (\ie early fusion), middle-level (halfway fusion), high-level (late fusion), and confidence-level (score fusion).
The first three approaches implement convolutional feature fusions, while the last corresponds to the combination of confidence scores from RGB and thermal \ac{CNN} branches at the decision stage.
The study reveals that the halfway fusion model achieves the lowest overall \ac{MR}.

In halfway fusion spatial attention-based mechanisms, the importance of each location in the feature map is calculated to highlight the areas with valuable information.
Accordingly, Cao \etal \cite{cao2023multimodal} used \ac{CSSA} in a lightweight fusion module to effectively fuse multi-modal inputs while ensuring low computational cost.
During channel switching, the channel of each modality with insufficient features is replaced by the corresponding channel from another modality.
Likewise, a bi-directional fusion strategy called \textit{BAANet} is introduced in \cite{yang2022baanet} to ensemble the RGB-thermal features for multi-spectral pedestrian detectors.
The strategy distills the high-quality features of two modalities and re-calibrates the representations gradually.
It contains intra- and inter-modality attention modules to improve spectra-specific features and adaptive selection of information from the most reliable modalities, respectively.
In another similar work, Zhang \etal \cite{zhang2021guided} introduced a two-stream \ac{CNN}, titled \acf{GAFF}, to dynamically re-weigh and integrate multi-spectral pedestrian features under the guidance of the intra- and inter-modality attention mechanisms.
The intra-modality attention module aims to enhance the visible or thermal features in pedestrian areas, while the inter-modality attention module selects the most reliable modality according to the feature quality, which requires costly annotation information.
The authors' solution to this issue is to assign labels based on the prediction of pedestrian masks from the intra-modality attention module and then select the most relevant modality where the prediction mask is closer to the ground truth.
Qingyun \etal \cite{qingyun2021cross} proposed a \ac{CFT} module and embedded it to the \ac{YOLO}~v5 framework.
The \ac{CFT} learns long-range dependencies and focuses on global contextual information.
In particular, by leveraging the self-attention mechanism, the network can simultaneously carry out intra-modal and inter-modal fusion and capture the latent interactions between visible and \ac{TIR} spectrums.

Some works discuss the most common feature fusion strategies in \acp{CNN}: \textit{concatenation} (\ie stacking two feature maps at the exact spatial locations), \textit{summation} (\ie calculating the sum of two feature maps at the exact spatial locations), \textit{maximum} (\ie obtaining the maximum response of two feature maps at the exact spatial locations), and \textit{mean} (\ie calculating the mean value of two feature maps at the exact spatial location) \cite{pei2020fast}.
Accordingly, Pei \etal \cite{pei2020fast} discussed the influence of these strategies in various \ac{CNN} fusion architectures, including merged \ac{VIS} and \ac{IR} images based on \textit{RetinaNet} detector \cite{lin2017focal}.
The results proved that the summation fusion strategy performs better than other methodologies.
Ding \etal \cite{ding2020convolutional} employed a \ac{NIN} in \ac{R-FCN} framework \cite{dai2016r} to merge the image information of two sub-networks to deal with large-scale and small-scale pedestrian instances.
After the concatenation of Conv-\ac{VIS} and Conv-\ac{IR}, the small- and large-scale pedestrian candidates generated by \ac{RPN} are merged with convolutional layers in the middle of the architecture.
Yun \etal \cite{yun2022infusion} proposed inter- and intra-weighted cross-fusion networks (Infusion-Net), which use a \ac{HFA} to integrate color and thermal features regarding the feature level gradually.
In this procedure, the \ac{HFA} block exchanges, purifies, and reinforces the object detection-relevant features based on \ac{DCT} and \ac{RCAB}.
Additionally, learnable inter- and intra-weight parameters provide optimal information utilization and feature reinforcement for each stream considering each fusion stage.
Bao \etal \cite{bao2023dual} proposed a dual-\ac{YOLO} method based on \ac{YOLO}~v7 \cite{wang2023yolov7} for integration of \ac{IR} and visible images.
They also designed attention fusion and fusion shuffle modules to alleviate the false detection rate caused by redundant feature information during the fusion process.

Numerous anchor-free pipelines have recently been proposed for multi-spectral pedestrian detection, which speeds up model detection while avoiding the complex hyper-parameter settings of anchor boxes.
Two feature fusion schemes based on a dual‐branch \textit{CenterNet} \cite{zhou1904objects} anchor-free detector proposed by \cite{zuo2022improving} for multi‐spectral and multi‐scale pedestrian detection.
The first one is \ac{SPA} module, which combines local attention and global attention sub‐modules, enhancing the quality of the feature fusion at different scales.
The second is \ac{AFA}, which aggregates features across different scales, considering spatial resolution and semantic context.
Likewise, Cao \etal \cite{cao2019box} attempted to train a multi-spectral pedestrian detector without anchor boxes via a box-level segmentation supervised learning framework and compute heat maps.
Consequently, the network can be able to localize the pedestrians on small-size input images.

In an innovative approach, Tang \etal \cite{tang2022piafusion} took illumination into account and designed a progressive image fusion network referred to as \textit{PIAFusion}, which can adaptively maintain the intensity distribution of salient targets as well as retain texture information in the background.
It uses an illumination-aware sub-network to estimate the illumination situations and exploits the illumination probability to construct illumination-aware loss.
Afterward, the \ac{CMDAF} module and halfway fusion strategy merge meaningful information of \ac{IR} and visible images under the guidance of illumination-aware loss.
Likewise, Roszyk \etal \cite{roszyk2022adopting} applied \ac{YOLO}~v4 framework for fast and low-latency multi-spectral pedestrian detection in autonomous driving.
Different fusion schemes, as well as different types of models, were investigated, among which feature-level fusion, namely \ac{YOLO}~v4-Middle, demonstrates the best trade-off between accuracy and speed.
Peng \etal \cite{peng2023hafnet} introduced \ac{HAFNet} embedded with a \ac{HCAF} module and a \acf{MFA} block to overcome the background noise and modality misalignment issue.
The \ac{MFA} exploits the correlation between the \ac{TIR} and visible domains to fine-tune the pixel alignment of multi-spectral image pairs.
Then, the \ac{HCAF} utilizes top-level features to guide pixel-wise fusion across two streams, resulting in high-quality feature representation.
Yadav \etal \cite{yadav2020cnn} built two uni-modal encoded-decoder feature networks for color and thermal individually using \ac{Faster R-CNN} \cite{ren2015faster}.
Further, they constructed middle-level \ac{CNN} fusion architecture, which fused the extracted features in the last convolution layer before feeding it to the decoder for providing the final predictions.
Zhang \etal \cite{zhang2019cross} presented a \ac{CIAN} to encode the correlations between two color and thermal spectrums and predict the positions and sizes of pedestrians on a contextual enhanced feature hierarchy.
Regarding the halfway fusion strategy, \ac{CIAN} has investigated three types of operations (\ie \textit{Elementwise Sum}, \textit{Elementwise Maximation}, and \textit{Concatenation and Channel Reduction}) for how to fuse feature maps.
The fusion operation of the \textit{Concatenation and Channel Reduction} shows better performance, which first concatenates the two feature maps, then applies a \ac{NIN} to reduce the number of channels.
To improve the detection accuracy in cases such as occluded objects, light changes, and cluttered backgrounds, Hu \etal \cite{hu2022dmffnet} proposed a \ac{DMFFNet}.
In their work, MobileNetv3 \cite{howard2019searching} extracts multi-scale features of dual-modal images as input for \ac{MFA} module, which processes the spatial information of input feature maps with different scales and establishes longer-distance channel dependencies, thereby reducing background noise interference.
Eventually, the \ac{DDFF} module deeply combines the multi-scale features to maximize the correlation between the multi-scale features, which significantly enhances the representation of semantic information and geometric detail.

Cao \etal \cite{cao2021attention} modeled a \ac{MCFF} module based on \ac{YOLO}~v4 to fuse the multi-spectral features according to the different illumination conditions.
The \ac{MCFF} module first concatenates the features from visible and thermal modalities in the channel dimension, then uses learning weights to adapt aggregate features.

\ac{GFU} \cite{kim2018robust} adjusts the contribution of the feature maps generated by each modality via the gating weighting mechanism.
Instead of stacking selected features from each channel and adjusting their weights, and motivated by \cite{kim2018robust}, \ac{GFD-SSD} \cite{zheng2019gfd} developed two variations of \ac{GFU} (\ie \textit{gated fusion} and \textit{mixed fusion}) to fuse the feature maps generated by the two \ac{SSD} \cite{liu2016ssd} middle layers for multi-spectral pedestrian detection.
Using \acp{GFU} on the feature pyramid structure, the authors also designed four mixed architectures of both stack fusion and gated fusion (\ie \textit{Mixed Even}, \textit{Mixed Odd}, \textit{Mixed Early}, and \textit{Mixed Late}), depending on which layers are selected to use the \acp{GFU}.
By comparing the experimental results on the \ac{KAIST} dataset, both the \ac{GFD-SSD} and \textit{Mixed Early} models are superior to the stack fusion.
\ac{RISNet} \cite{wang2022improving} designed a mutual information minimization module to alleviate the influence of cross-modality redundant information on the fusion of RGB-\acl{IR} complementary information.
Besides, the \ac{RISNet} introduced a classification method of illumination conditions based on histogram statistics.

\noindent\textbf{Late fusion-based methods:}
also known as decision-level fusion, is the high-level fusion technique in which the concatenation is conducted after the last convolutional layer and before fully connected layers or merged the outputs of the two sub-networks such as the location and category prediction.
As a work in this category, \ac{MS-DETR} is introduced by \cite{xing2023multispectral}, which extracts multi-scale feature maps through two parallel modality-specific \ac{CNN} backbones, aggregates them within the corresponding modality-specific transformer encoders, and fuses the features using a multi-modal transformer decoder.
It also adopts a modality-balanced optimization strategy to measure further and balance the contribution of each modality at the instance level.
Khalid \etal \cite{khalid2019person} proposed two fusion methods to detect people:
In the first one, an encoder-decoder architecture was used for image-level fusion, which independently encodes visible and thermal frames and fed the combined frames into a decoder to produce a single fused image, inputting a \ac{ResNet-152} architecture.
The second one takes \ac{ResNet-152} for feature-level fusion, which extracts features of visible and thermal images separately and concatenates them into a single feature vector as the input of the dense layer.
Montenegro \etal \cite{montenegro2022pedestrian} customized \ac{YOLO}~v5's architecture for low-light pedestrian detection, and also they have conducted experiments on multiple multi-spectral pedestrian datasets \eg CVC-09, \ac{LSI-FIR}, \ac{FLIR}, CVC-14, NightOwls, and \ac{KAIST}.
By extensive evaluations made on different datasets, the best \ac{mAP} was obtained on \ac{LSI-FIR}, followed by CVC-09 and CVC-14.
Song \etal \cite{song2021multispectral} designed a \ac{MSFFN} that uses the extracted features of visible-channel and infrared-channel to obtain integrated features.
The \ac{MSFFN} strikes a favorable trade-off between accuracy and speed, especially on small-size input images.
They extract multi-scale semantic features using two sub-networks, including \ac{MFEV} and \ac{MFEI} and integrate them with an improved \ac{YOLO}~v3 framework.
\ac{SKNet} \cite{ding2021robust} suggested a dynamic selection scheme to adaptively adjust receptive field size using selective kernel units with different kernel sizes.
It uses a \ac{NIN}-based fusion strategy to fuse RGB-\ac{IR} image pairs.
Park \etal \cite{park2018unified} considered all detection probabilities from RGB, \ac{IR}, and RGB-\ac{IR} fusion channels in a unified three-branch model and designed a \ac{CWF} and an \ac{APF} layers to fuse probabilities from different information streams at a proposal-level.
A combination of an adaptive weight adjustment method with the \ac{YOLO}~v4 \cite{bochkovskiy2020yolov4} is introduced by \cite{chan2023multispectral} to enrich the multi-spectral complementarity information for score fusion.
Authors in \cite{chen2022multimodal} introduced \ac{ProbEn}, a simple non-learning technique for late-fusion of multiple modalities derived from Bayes' first principle, \ie conditional independence assumptions.
Shaikh \etal \cite{shaikh2022probabilistic} introduced a probabilistic decision-level fusion approach based on Naïve Bayes to address lighting and temperature changes in color and thermal images by fusion and modeling the detection results of available pedestrian detectors without requiring retraining.
In particular, the use of Naïve Bayes for the late fusion strategy enables the network to work with non-registered image pairs as well as poorly registered image pairs.

Zhuang \etal \cite{zhuang2021illumination} examined the impacts of environmental variables on the efficiency of the pedestrian detector and proposed a lightweight \ac{IT-MN}.
The method is built on the \ac{SSD} architecture with designing a late-fusion strategy and \ac{FWN} to compute the fusion weights.
In addition, the default box generation is optimized by reducing the number of bounding boxes and choosing specific box aspect ratios to minimize the inference time.
Inspired by \ac{YOLO}~v4, \ac{DSMN} \cite{hsia2023all} was designed to carry out pedestrian detection in challenging situations such as insufficient and confusing lighting.
Their method extracts multi-spectral information provided by RGB and thermal images via two \ac{YOLO}-based sub-networks.
Also, it has an \ac{i-IAN} module to estimate the lighting intensity of varied scenarios and allocate fusion weights to RGB-thermal sub-networks.
Li \etal \cite{li2019illumination} explored various fusion schemes and pointed out their key adaptations.
They also designed an \ac{IAF R-CNN} framework to estimate the illumination value of the input image and incorporate color and thermal sub-networks via a gate function defined over the illumination value.
In another work, Li \etal \cite{li2022pedestrian} introduced an \ac{ASG-LPF} to improve detection performance in varying lighting conditions, which uses a light perception module to distinguish the illumination levels in diverse driving scenarios.
Takumi \etal \cite{takumi2017multispectral} proposed a multi-spectral ensemble method based on \ac{YOLO}~v1 \cite{redmon2016you}, which integrates detection results of the four single-spectral detection models into a single space as the final detection.
As another approach, LG-FAPF \cite{cao2022locality} performed a cross-modal feature aggregation process guided by locality information to learn human-related multi-spectral features and used the obtained spatial locality maps of pedestrians as pixel-wise prediction confidence scores for the adaptive fusion of detection results under complex illumination conditions.
Considering \ac{CSPNet} model, Wolpert \etal \cite{wolpert2020anchor} proposed an anchor‐free multi‐spectral framework to investigate various fusion strategies.
They also introduced a new data augmentation technique for multi-spectral images called \textit{Random Masking}.
\subsubsection{\textbf{Knowledge Transfer-based Methodologies}}
\label{subsubsec_knowledge-transfer}

This section's approaches leverage insights from various domains based on knowledge acquired from diverse sources to facilitate nighttime pedestrian detection capabilities.
It contains various categories with different methodologies, including transfer learning, supervised and unsupervised domain adaptation, knowledge distillation, and memory-network methods.

\noindent\textbf{Transfer learning methods:}
transfer learning is reusing the knowledge obtained from pre-trained models for dissimilar but related tasks.
In the context of nighttime pedestrian detection, and to fill in the gap of large-scale \ac{TIR} dataset, Hu \etal \cite{hu2020application} applied CycleGAN \cite{zhu2017unpaired} to generate synthetic \ac{IR} images from visible ones to expand the \textit{CVC-09} dataset.
They performed experiments using the \ac{YOLO}~v3 and \ac{Faster R-CNN} models on the \textit{CVC-09} dataset, in which the \ac{Faster R-CNN} has shown better performance in the transfer learning task.
In another work by Vandersteegen \etal \cite{vandersteegen2018real}, a pre-trained \ac{YOLO}~v2 \cite{redmon2017yolo9000} was used to perform real-time visible-thermal pedestrian detection.
Their method takes three image channels composed of a combination of four image channels (\ie \textit{RGB} and \textit{T}) information as input and can work on 80 \ac{fps}.
They discussed the possibility of creating a number of channel combinations as input channels of the \ac{YOLO}~v2 model and designed three models named \ac{YOLO}-TGB, \ac{YOLO}-RTB and \ac{YOLO}-RGT.
The \ac{YOLO}-TGB, which only uses the combination of thermal, green, and blue image channels as input, performs better on the \ac{KAIST} dataset than other proposed models.
Geng \etal \cite{geng2020infrared} replaced the loss function of \ac{YOLO}~v3 model with \textit{DIOU Loss} \cite{zheng2020distance} to accelerate the convergence speed of the network in \ac{IR} image-based pedestrian detection.
Although the loss function curve is more stable, the \ac{AP} of the Diou-\ac{YOLO}~v3 is not satisfactory.

\noindent\textbf{Domain adaptation methods:}
the critical idea of employing the domain adaptation mechanism in multi-spectral pedestrian detection is to exploit learned knowledge acquired from the color domain in the thermal images.
In this regard, Guo \etal \cite{guo2019domain} focused on image-level domain adaptation by using an image-to-image transformer as a data augmentation tool to convert color images to the thermal spectrum.
To aid the joint training process of the domain adapter and the detector, the authors defined a detection loss that back-propagates its gradients to the image transformer to progressively refine synthetic thermal images.
The proposed method provides promising results compared to the baseline on the \ac{KAIST} benchmark.
Kieu \etal \cite{kieu2020task} introduced a task-conditioned training method to help domain adaptation of \ac{YOLO}~v3 to the thermal spectrum.
The primary detection network was augmented by adding an auxiliary classification task of day and nighttime thermal images.
Additionally, learned representations of this auxiliary task were used to condition \ac{YOLO} to perform better in the thermal imagery.
Authors in \cite{kieu2019domain} addressed three top-down and one bottom-up domain adaptation techniques for pedestrian detection in the nighttime thermal images.
They showed that bottom-up domain adaptation achieves better results in challenging illumination conditions.
As another work by the same authors \cite{kieu2021bottom}, a new bottom-up domain adaptation strategy, known as \textit{layer-wise domain adaptation}, is introduced.
The main idea for this method is to adjust the RGB-trained detector to adapt to the thermal spectra gradually.
Kristo \etal \cite{krivsto2020thermal} attempted to improve the typical object detector performance for person detection at night in challenging weather conditions such as heavy rain, clear weather, and fog.
The authors retrained \ac{YOLO}~v3, \ac{SSD}, \ac{Faster R-CNN}, and Cascade R-CNN detectors on a dataset of thermal images.
They found that \ac{YOLO}~v3 is significantly faster than the others with a processing speed of $27,5$ \ac{fps}.
The generalization ability of \ac{RPN} has been analyzed by \cite{fritz2019generalization} for multi-spectral person detection by performing cross-dataset evaluations on several benchmark datasets such as Caltech \cite{dollar2011pedestrian}, CityPersons \cite{zhang2017citypersons}, CVC-09, \ac{KAIST}, OSU, and Tokyo semantic segmentation \cite{ha2017mfnet}.
They showed that \ac{KAIST} achieves better results in generalization tasks in both daytime and nighttime conditions.

\noindent\textbf{Unsupervised domain-adaptation methods:}
the objective of unsupervised domain adaptation is to adapt the well-trained detectors on annotated visible images to the thermal target without any manual annotation effort.
As a work in this category, Meta-UDA \cite{vs2022meta} performed \ac{UDA} thermal target detection using an online meta-learning strategy, resulting in a short and tractable computational graph.
To mitigate the domain shift between the source and target domain, the Meta-UDA uses the adversarial feature alignment at both the image and instance levels, leading to slight improvement.
In another work, Lyu \etal \cite{lyu2021visible} used an iterative process to automatically generate the pseudo-training labels from visible and thermal modalities using two single-modality auxiliary detectors.
They used the illumination knowledge of daytime and nighttime to assign the fusion priorities of labels for \textit{label fusion}.
Without using any manual training labels on the target dataset, the proposed method shows reasonable results on the night scenes of the \ac{KAIST} dataset.
Authors in \cite{munir2021sstn} used transformers to tackle unlabeled data challenges in \ac{TIR} images.
They designed a \ac{SSTN} to learn feature representation and maximize the mutual data between visible and \ac{IR} domains by contrastive learning to compensate for the shortage of labeled data.
Later, a multi-scale encoder-decoder transformer system was employed for thermal object detection based on the learned feature representations.
Inspired by pseudo-training labels, Lyu \etal \cite{lyu2022unsupervised} proposed an unsupervised transfer learning framework in multi-spectral pedestrian detection.
Their overall framework is based on a two-step domain adaptation solution, in which the first stage generates intermediate representations of color and thermal images to reduce the domain gap across the source and target domains.
The pseudo labels of the target objects are fused via an illumination-aware label fusion mechanism.
In the second stage, an iterative fine-tuning process is conducted to progressively converge the detector on the target domain.
In another work, Cao \etal \cite{cao2019pedestrian} introduced an auto-annotation framework to iteratively label pedestrian instances in visible and thermal image channels by leveraging the complementary information of multi-modal data.
They aim to automatically adapt a pedestrian detector pre-trained on the visible domain to a new multi-spectral domain without manual annotation.
The predicted pedestrian labels on both image channels are merged via a label fusion scheme to generate the final multi-spectral pedestrian annotations.
Then, the automatically generated labels are fed to a \ac{TS-RPN} detector to achieve unsupervised learning of complementary semantic features.
An unsupervised multi-spectral domain adaptation framework was proposed by Guan \etal \cite{guan2019unsupervised} to generate pseudo-annotations in the source domain, which can be utilized to update the parameters of the model in the target domain according to the complementary information in aligned visible-\ac{IR} image pairs.
Transfer knowledge from thermal to visible domain in unpaired settings and without requiring additional annotations has been performed in \cite{marnissi2022unsupervised} by applying image-level and instance-level alignments based on the \ac{Faster R-CNN} network using adversarial training.

\noindent\textbf{Knowledge distillation methods:}
The concept of the \ac{KD} is based on inheriting the knowledge learned from a large and complex pre-trained teacher model to a smaller and simpler student model through a supervised learning process \cite{hinton2015distilling}, \cite{hnewa2023cross}, \cite{zhang2022low}.
Generally, the main objective of this method is to transfer the applicable and meaningful representations of data to speed up the inference time of the student model without a significant drop in accuracy \cite{hnewa2023cross}.
According to the teacher-student scheme, Liu \etal \cite{liu2021deep} developed a knowledge distillation framework as a student network that only takes color images as input and generates distinguishing multi-spectral representations, guided by a two-modalities teacher network.
Moreover, \acf{CFL} module based on a split-and-aggregation approach was incorporated into the teacher network to learn the standard and modality-specific characteristics between color and thermal image pairs.
Hnewa \etal \cite{hnewa2023cross} employed \ac{CMKD} to enhance the performance of RGB-based pedestrian detection under adverse weather and low-light conditions.
Two different \ac{CMKD} methods were developed to transfer the multi-modal information of a teacher detector to a student RGB-only detector.
The former uses \ac{KD} loss, while the latter integrates adversarial training with knowledge distillation.
Zhang \etal \cite{zhang2022low} proposed a \ac{MD} framework to transfer the knowledge from a high thermal resolution two-stream network with feature-level fusion to a low thermal resolution single-stream network with early fusion strategy.
In particular, two specific knowledge distillation modules are used in the \ac{MD} framework.
An attention transfer generates attention masks by \ac{GAFF} from a two-stream teacher model, which is transferred to a single-stream student model through performing an early fusion.
Finally, a semantic transfer resolves the problem of modality imbalance in feature distillation using a new \ac{F-MSE} cost function.

\noindent\textbf{Memory-network methods:}
\ac{MANN} can memorize and recall the prior information, such as visual appearance in the memory module, so the relevant data can be accessed by calculating the similarity \cite{kim2021robust}.
In \cite{kim2022towards}, a pedestrian detection process is introduced to improve the detector's performance in any modality.
In the first stage, a multisensory-matching contrastive loss guides the pedestrian visual representation of two visible and thermal modalities to be similar.
In the second, a \ac{MSR} memory improves the visual representation of the single modality features by recalling the visual appearance of multi-spectral modalities and memorizes the multi-spectral contexts through a multi-spectral recalling loss, which encoded more discriminative information from a single input modality.
The \acf{LPR} based on key-value memory was proposed by \cite{kim2021robust}, which memorizes visual information of large-scale pedestrians to recall the relevant characteristics to cover inadequate small-scale pedestrian appearances.
\subsubsection{\textbf{Image Specification-based Methodologies}}
\label{subsubsec_image_specs}

Another category focuses on methodologies in which image specifications play a crucial role.
These methods can be divided into three primary strategies: \textit{image enhancement}, \textit{image-to-image translation}, and \textit{saliency maps} methods.

\noindent\textbf{Image enhancement methods:}
\ac{TIR} images are characterized by noisy details, blurred edges, low contrast, and low resolution, resulting in a performance drop caused by low discrimination.
In this regard, low-light image enhancement techniques are considered to improve the visual quality of thermal images and simplify their challenges.
In a work by Marnissi \etal \cite{marnissi2021thermal}, an enhancement method based on images' architecture, title \acf{TE-GAN} is designed, which constituted of contrast augmentation, noise elimination, and edge restoration.
To enhance the clarity of the \ac{IR} pedestrian targets with blurred edges, Sun \etal \cite{sun2021method} adopted a super-resolution algorithm called \ac{WDSR}-B \cite{yu2018wide}.
They add the four-time down-sampling layer output to \ac{YOLO}~v3 trained by the enhanced \ac{IR} images to acquire richer context information for small pedestrian targets.
In another work, Marnissi \etal \cite{marnissi2023gan} combined \ac{GAN} and \ac{ViT} for thermal image enhancement and introduced \ac{TE-VGAN}.
\ac{TE-VGAN} employs the U-Net architecture as an input image generator and two \ac{ViT} models as global and local discriminators.
The thermal loss feature is also introduced in their work to generate high-quality images.
They investigated the effect of the thermal image enhancement method on the detection performance of different \ac{YOLO} versions, resulting in a balance between contrast enhancement and noise reduction.

DIVFusion \cite{tang2023divfusion} incorporates a low-light image enhancement task and a dual-modal fusion task in a unified framework to investigate the effect of lighting conditions on image fusion.
In their method, firstly, a \ac{SIDNet} is devised to eliminate the illumination degradation in nighttime visible images while maintaining informative features of source images.
Then, a \ac{TCEFNet} is employed to aggregate complementary information and boost fused features' contrast and texture details.
Finally, a color consistency loss is used to alleviate color distortion in enhancement and fusion processes.
Li \etal \cite{li2021nighttime} built \ac{FAM} and \ac{FTM} to improve the efficiency of a pedestrian detector in darkness.
\ac{FAM} is designed to suppress the noisy representations, while \ac{FTM} allows pedestrian examples under a low-light environment to generate more discriminate feature representations.
An attention-based feature fusion module was designed in \cite{cui2023bright} to enhance pedestrian detection in low-illumination images.
They used the brightness channel (\ie V-channel) from the \textit{HSV} image of the thermal image as an attention map to activate the unsupervised auto-encoder for obtaining more details about the pedestrian.
In order to address the challenge of light compensation in low-light conditions, a Brightness Correction Processing (BCP) algorithm is considered to guide self-attention map learning.
Eventually, the image enhancement method was integrated into \ac{YOLO}~v4 detection model.
They evaluated the proposed architecture on the \ac{LLVIP} dataset.

To highlight pedestrians in low-resolution and noisy \ac{IR} images, an \ac{AED-CNN} \cite{chen2020pedestrian} is devised.
In \ac{AED-CNN}, the encoder-decoder module generates multi-scale features, and a skip connection block is integrated into the decoder to fuse the feature maps from the encoder and decoder structure.
By adding an attention module, the network effectively emphasizes informative features and suppresses background interference while re-weighting the multi-scale features generated by the encoder-decoder module.
Patel \etal \cite{patel2022depthwise} introduced a computationally compact algorithm based on Depthwise Convolution (DC) with the aim of network parameters reduction.
The proposed algorithm enhances the details of the thermal images using \ac{AHE} and extracts the salient features in these images by a new \ac{CBN}, where depthwise convolution minimizes the computational complexity.
YOLO-FIRI \cite{li2021yolo} is another method developed for pedestrian detection in \ac{IR} images, which achieved outstanding results by making improvements on \ac{YOLO}~v5 structure.
Firstly, by extending shallow \ac{CSPNet} in the backbone network and incorporating an improved \ac{SK} attention module in the residual block, it forces the model to focus on shallow and detailed information and learn the distinguishable features.
Secondly, the detection accuracy of small and blurry pedestrians in \ac{IR} images is increased by adding four-scale feature maps to the detection head.
Finally, Densefuse \cite{li2018densefuse} is adopted as a data enhancement to fuse visible and infrared images to boost the features of \ac{IR} images.

\noindent\textbf{Image-to-image translation methods:}
the goal of \ac{I2I} translation models is to learn the visual mapping between a source and target domain while preserving the essential features.
Specifically, \ac{I2I} has been widely used in image colorization, denoising, and synthesis \cite{pang2021image}.
In these approaches, thermal image colorization aims to translate from the temperature-channel domain into the RGB channel.
PearlGAN presented in \cite{luo2022thermal} to facilitate the translation of nighttime \acf{TIR} image into a daytime color one.
By taking advantage of a top-down guided attention module and a corresponding attention loss, \textit{PearlGAN} can produce hierarchical attention distribution and reduce local semantic ambiguity in \ac{IR} images through context information.
In addition, a structured gradient alignment loss was designed to enhance edge consistency during the translation.
The colorization of thermal-\ac{IR} images in pedestrian detection application is accomplished by \cite{dangle2023enhanced}, organized into three main modules: thermal image colorization, improvement of colorized images, and pedestrian detection.
The colorized and improved images are fed to the detection head using a pre-trained \ac{YOLO}~v5 framework.

To mitigate color distortion and edge blurring caused by translation from temperature spectrum to color spectrum, \cite{yang2023unpaired} considered a one-to-one mapping relationship and introduced an improved CycleGAN \cite{zhu2017unpaired}, called \ac{GMA-CycleGAN}.
It first translates the \ac{TIR} images to \ac{GV} and then uses the original CycleGAN to obtain the translation from \ac{GV} to \ac{CV}.
A mask attention module based on the thermal temperature mask and the color semantic mask has been designed without increasing training parameters to better differentiate between pedestrians and the background.
Meanwhile, to make the texture and color of the translated image more realistic in the feature space, a perceptual loss was added to the original CycleGAN loss function.
Devaguptapu \etal \cite{devaguptapu2019borrow} proposed to borrow knowledge from the large-scale RGB dataset without the need for paired multi-modal training examples and used CycleGAN to implement an unpaired image-to-image translation framework.
It can generate pseudo-RGB equivalents of a given thermal image and employs a multi-modal \ac{Faster R-CNN} detector for pedestrian detection in thermal imagery.
To transform the visible domain into the thermal domain, authors in \cite{kieu2021robust} implemented a generative data augmentation method based on the \acf{LS-GAN} \cite{mao2016multi}.
They also used the perceptual loss function to measure the similarity between authentic and synthesized images in pixel space.

\noindent\textbf{Saliency maps methods:}
the purpose of salient object detection is to highlight the most noticeable areas in the given image and distinct the prominent objects from their surroundings using the intensity of each pixel.
Accordingly, in \ac{TIR} images, the saliency maps can be used to detect temperature.
Altay \etal \cite{altay2022use} presented a two-branch architecture that can incorporate features of thermal images with their correlated saliency maps to acquire better representations of pedestrian regions.
Instead of using color-thermal image pairs in the fusion network, Ghose \etal \cite{ghose2019pedestrian} suggested augmenting thermal images with their corresponding saliency maps, which produced by static methods and two deep saliency networks, \ac{PiCANet} \cite{liu2018picanet} and \ac{RRRNet} \cite{deng2018r3net}.
Marnissi \etal \cite{marnissi2022bispectral} proposed a bi-spectral image fusion scheme, which was augmented with a corresponding saliency map using \ac{VST} and also incorporated this fusion process into the \ac{YOLO}~v3 as base architecture for real-time applications.
The proposed approach has shown its advantage in low computational cost, which allows faster inference time.
Zhao \etal \cite{zhao2019infrared} put more emphasis on the temperature information in infrared images by constructing an \ac{IR}-temperature transformation formula which can convert the \ac{IR} images into corresponding temperature maps.
It finally uses a trained temperature network for pedestrian detection.
On the \ac{OSU} and \ac{FLIR} datasets, the transformed temperature maps boost the overall performance regardless of external influences.
\subsubsection{\textbf{Images Discrepancy-based Methodologies}}
\label{subsubsec_image_discrepancy}

These methods target enhancing the accuracy and reliability of nighttime pedestrian detection by exploiting discrepancies within images and characteristics of different imaging sensors and analyzing variations in image quality and content.
The modality discrepancy is alleviated by focusing on \textit{modality imbalance problem} and \textit{position-shift problem}.

\noindent\textbf{Modality imbalance problem:}
Scenes in which one sensor performs considerably better than the others can lead to a \textit{bias} in training towards one dominant input modality.
For instance, uneven distribution of training data in multi-modal learning causes less contribution of the non-dominant input modality during network training and, therefore, limits the generalizability of the model.
In this regard, Oksuz \etal \cite{oksuz2020imbalance} provided a comprehensive taxonomy of the imbalance problems in object detection.
They categorize these problems into four significant categories: \textit{class imbalance} (\ie inequality distribution of training data among different classes), \textit{scale imbalance} (\ie various scales of objects), \textit{spatial imbalance} (\ie spatial properties of the bounding boxes), and \textit{objective imbalance} (\ie minimization of multiple loss functions).
Additionally, in multi-spectral pedestrian detection, the modality imbalance issue substantially impacts the algorithm performance, which can occur in two different ways, including the \textit{illumination modality imbalance problem} and the \textit{feature modality imbalance problem} \cite{zhou2020improving}.
Das \etal \cite{das2023revisiting} proposed a training process with a regularization term \ie \textit{Logarithmic Sobolev Inequalities} \cite{gross1975logarithmic} to consider the features of both modalities equally during fusion.
The proposed regularizer reduces the modality imbalance in the network by equally distributing the training data among the modalities.
Li \cite{li2021infrared} trained \ac{YOLO}~v3 framework to detect pedestrians under insufficient illumination conditions.
In their method, focal loss \cite{lin2017focal} was added to the loss function to overcome the imbalance issue of \ac{IR} images.
Zhou \etal \cite{zhou2020improving} resolved the modality imbalance issue in multi-spectral images through the implementation of a single-stage \ac{MB-Net}, which included a \ac{DMAF} and an \ac{IAFA} module to extract complementary information and align the two modality features according to the lighting conditions.
Dasgupta \etal \cite{dasgupta2022spatio} developed \acf{MuFEm} module using \acf{GAT} \cite{velickovic2017graph} to deal with the imbalance issue between color image branch and thermal image branch.
Also, the channel-wise attention block and four-directional IRNN (4Dir-IRNN) block \cite{bell2016inside} are incorporated in \acf{SCoFA} to improve fusion using spatial and contextual information of the pedestrian.
The \textit{4Dir-IRNN} block consists of four \acfp{RNN}, which compute context features in four directions.

\noindent\textbf{Position-shift problem:}
The physical properties of different cameras (\eg \acf{FoV}, resolutions, wavelengths, \etc) can cause weakly aligned image pairs in multi-spectral data, where the positions of the objects are out of synchronization on different modalities.
Some works tried to address the mentioned problem in multi-modal sensors using geometrical calibration and image alignment methods.
The study by Zhang \etal \cite{zhang2019weakly} is the first work providing insights into the position shift problem between color and thermal images.
They introduced an \ac{AR-CNN} detection framework to solve the weakly aligned image pairs.
The \ac{AR-CNN} firstly predicts the position shift and adaptively aligns the region feature maps of the two modalities through a \ac{RFA} module.
Based on the aligned features, a confidence-aware fusion method is proposed to accomplish feature re-weighting, which selects the highly informative features while suppressing the useless ones.
Moreover, a \ac{ROI} jitter strategy is adopted to enhance the robustness of position shift patterns.
Kim \etal \cite{kim2019unpaired} used adversarial learning to make each spectrum share its complementary information in a common feature space to compensate for the lack of aligned multi-spectral pedestrian datasets.
Kim \etal \cite{kim2021uncertainty} have constructed uncertainty-aware multi-spectral pedestrian detection architecture to handle miscalibration (\ie different \ac{FoV} in color and thermal cameras) and modality discrepancy challenges.
For the miscalibration issue, the \acf{UFF} module was formulated to mitigate the impact of ambiguous \acf{ROI}.
The modality discrepancy is alleviated through the \acf{UCG} module, which can encode more discriminative visual representations.
Wanchaitanawong \etal \cite{wanchaitanawong2021multi} introduced a multi-modal \ac{Faster R-CNN} robustly against significant misalignment between the two modalities.
The key points are modal-wise regression for bounding-box regression of each modality to deal with the significant misalignment and multi-modal \ac{IoU} for mini-batch sampling that combines the \ac{IoU} for both modalities.
\subsubsection{\textbf{Multi-task methods}}
\label{subsubsec_multi-task} 

Multi-task learning is a training paradigm that aims to learn multiple related tasks simultaneously, using shared feature representations \cite{crawshaw2020multi}.
A cross-task feature alignment method was proposed by \cite{wang2021cross} to tackle the misalignment of scale and channel of features from image relighting and pedestrian detection tasks by placing four feature alignment layers before the feature fusing and sharing step in cross-task learning.
Meanwhile, a multi-scale feature-enhanced detection network expands the receptive field of the multi-scale feature extractor and thereby provides richer semantic information of fused features for the detection head.
An illumination-aware weighting mechanism is presented by Guan \etal \cite{guan2019fusion} to adaptively re-weight the detection results of day- and night-illumination sub-networks to learn multi-spectral human-related characteristics to perform pedestrian detection and semantic segmentation under various illumination conditions, simultaneously.
Dai \etal \cite{dai2021multi} developed the \ac{Faster R-CNN} detector using the ResNet-$50$ as a feature extractor for pedestrian detection and distance estimation using the \ac{NIR}-based camera.
An \acf{ARPN} was designed by \cite{cao2019new} to get bounding boxes.
A pedestrian segmentation task is also added based on a \acf{FPN} \cite{lin2017feature} to obtain the confidence scores.
To distinguish pedestrian examples from complex negative samples, Li \etal \cite{li2018multispectral} added two sub-networks for jointly semantic segmentation and pedestrian detection tasks to the unified fusion network, which is denoted as \textit{MSDS-RCNN}.
The paper also studied the effects of training annotation noise by creating a sanitized version of \ac{KAIST} ground-truth annotations so that the sanitized training annotations significantly reduce the inference error.
Evaluations showed that the segmentation supervision benefits multi-spectral pedestrian detection.
\subsubsection{\textbf{Other methods}}
\label{subsubsec_other}

As for the final category, this subsection introduces works that cannot fit into the previous ones.
Accordingly, the authors in \cite{chen2019thermal} employed a region decomposition branch in \ac{Faster R-CNN} architecture, which exploits the multi-region features, including head, body trunk, and legs, to solve the pedestrian occlusion problem in thermal images.
The proposed architecture learns the high-level semantic features by combining the global and partial appearance features step by step.
The \acf{CSPNet} \cite{liu2019high} has been applied in \cite{song2020full} to obtain three \ac{IR} pedestrian detection models, namely daytime, nighttime, and full-time.
The full-time model has a lower detection loss rate, while the nighttime model and the daytime model perform poorly in detecting small objects in the evening, respectively.
Xu \etal \cite{xu2020ground} aggregated ground-area context information into the \ac{Faster R-CNN} for pedestrian detection and shared the predicted ground horizon area to a \acf{GRPN}, which can only process the pixels on the proposed horizon region to minimize \acf{FP} rate.
Since the output of the \textit{FC layer} is the position vector of pixels in the horizon region, the size of the \ac{GRPN} model is largely increased and has a high computational cost.
Dai \etal \cite{dai2019near} compared and analyzed visible and \ac{IR} images acquired by using visible-spectrum, \acf{NIR}, \ac{SWIR}, and \ac{LWIR} cameras.
For the first time, they used a nine-layer \ac{CNN} model with a self-learning SoftMax \cite{bishop2006pattern} to detect nighttime pedestrian samples in \ac{NIR} images.
In order to enhance the detection accuracy of multi-scale pedestrians in \ac{IR} images, in \cite{galarza2018pedestrian}, two regional proposal networks based on the \ac{Faster R-CNN} architecture were designed to focus on near and far away pedestrians.
Although the proposed multi-scale \ac{RPN} has shown improvements in far-away pedestrian detection, it is not optimized to work in real time.
Kalita \etal \cite{kalita2020real} have presented a real-time human detection system using \ac{YOLO}~v3, which achieved a speed of $17$ millisecond per image on the \ac{KAIST} thermal dataset.
The brightness aware \acf{Faster R-CNN} model \cite{chebrolu2019deep} was proposed to perform the pedestrian prediction under low-light and day-light scenarios.
In the first step, the model calculates the brightness of the input image based on the pixel intensity to predict the day or night scenario.
In the second step, two separate thermal or color models are employed for pedestrian detection based on the first step output.
It should be noted that the authors trained the \ac{FLIR} dataset for the thermal model and the PASCAL VOC dataset for the color model.
\subsection{Hybrid Approaches}
\label{subsec_hybrid}

Hybrid methods combine elements of handcrafted features and deep learning, aiming to harness the strengths of each approach for improved nighttime pedestrian detection performance.
Accordingly, they require significantly less computational resources than deep learning methods and overcome the poor generalization of handcrafted methods.
However, the performance of such approaches is not properly optimized in terms of prediction accuracy or model running time.

As a hybrid methodology for nighttime pedestrian detection, the study of Kim \etal \cite{kim2018pedestrian} presented a method to detect pedestrians at night using a visible-light camera and \ac{Faster R-CNN} model, which can handle the changes of the pedestrians' spatial position by fusing deep convolutional features in successive frames.
To make the model robust against noise and illumination, the authors used \ac{AWGN} and applied two pre-processing methods, \ie pixel normalization and \ac{HE} mean subtraction, to normalize the illumination and contrast levels of successive frames.
Besides, a weighted summation of successive frame features was added to exploit temporal information about the pedestrian, which enhanced the accuracy of the pedestrian detector at nighttime.
To find the optimal fusion stage in \ac{CNN}, authors in \cite{konig2017fully} used \ac{RPN} to merge the features of visual and \ac{IR} images.
After halfway feature fusion in \ac{RPN}, they employed \ac{BDT} classifier to improve pedestrian detection results and reduce the false positive rate.
Tumas \etal \cite{tumas2018acceleration} eliminated the sliding window technique and applied background subtraction to extract thermally active points as \acf{ROI} for pedestrian detection in \acf{FIR} domain.
The proposed technique accelerates the \acf{HOG} based pedestrian detector to run at $6$ \ac{fps} using only CPU performance.
Narayanan \etal \cite{narayanan2021study} developed a model for low-light pedestrian prediction using \ac{HOG} and \ac{YOLO}~v3 algorithm.
They also experimented the detection accuracy of \ac{HOG} detector and \ac{SVM} classifier in thermal images.
Xu \etal \cite{xu2017learning} designed a framework for learning and transferring cross-domain feature representations for pedestrian detection that works based on two different networks.
The first one, titled \ac{RRN}, is employed to learn a non-linear feature mapping and model the relations among the color and \ac{IR} image pairs.
Afterward, the cross-modality feature representations learned from \ac{RRN} are transferred to a second network titled \ac{MSDN}, which operates only on RGB inputs and outputs the recognition results.
Both \ac{RRN} and \ac{MSDN} networks have employed ACF \cite{dollar2014fast} to generate pedestrian proposals.
In this way, only color images are considered at the test phase, and no thermal data are needed, which significantly reduces the cost of thermal data annotation.
In \cite{choi2016multi}, \ac{SVR} was adopted to learn the pedestrians probabilities, which performs well on small-scale pedestrians.
Chen \etal \cite{chen2021robust} utilized a \ac{TV} minimization \cite{ma2016infrared} method based on structure transfer to integrate \ac{TIR}-RGB image pairs, preserving the infrared intensity distribution and the local appearance features.
However, when the thermal radiation of the pedestrian and the background are the same, the performance of the detector is affected.
\section{Discussion}
\label{sec_discussion}

Regarding the state-of-the-art methodologies introduced in previous sections, this section discusses the current trends and future expectations of the works targeting nighttime pedestrian detection.

\subsection{Employed Methodologies Trends}
\begin{figure}[t]
    \centering
    \includegraphics[width=1.0\columnwidth]{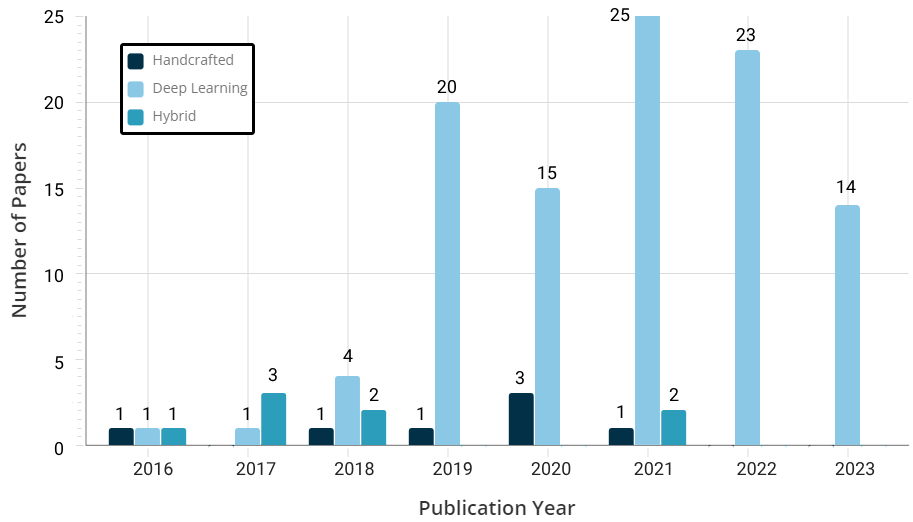}
    \caption{The trends of employing various approaches explained in this survey by state-of-the-art research works published in different years.}
    \label{fig_chart_approach}
\end{figure}

Regarding categorizing the works introduced in Section~\ref{sec_sota}, nighttime pedestrian detection approaches can be divided into handcrafted features, deep learning, and hybrid methodologies.
In this regard, Fig~\ref{fig_chart_approach} shows the distribution of the surveyed paper regarding the primary categories they belong to.
According to the figure, it can be seen that the majority of the works published in the last two years consider deep learning-based techniques the most reliable methodology to detect pedestrians in low-light conditions.
In other words, recent approaches only focus on employing \acp{DNN} instead of handcrafted and hybrid approaches.
The main reason may be attributed to \textit{automatic learning of features} in \acp{DNN}, which cover many possible conditions in which pedestrians are challenging to detect.
Additionally, the works are getting more practical, providing the possibility to be used in real applications, and making domain-specific applications based on handcrafted or hybrid methods is not a practical solution.

\begin{figure}[t]
    \centering
    \includegraphics[width=1.0\columnwidth]{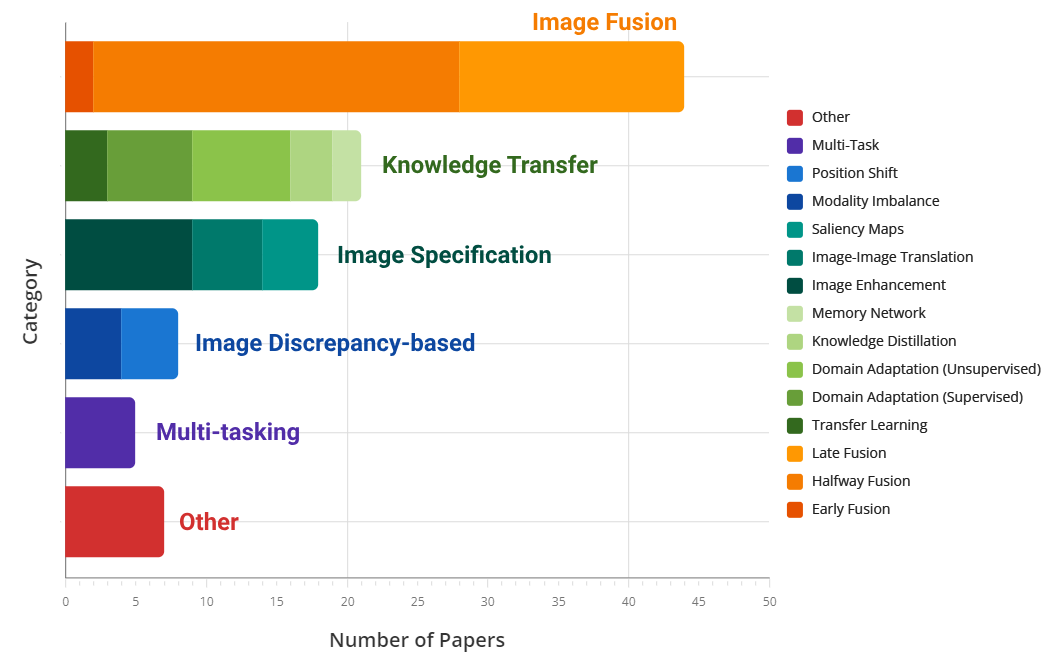}
    \caption{Distribution of the reviewed papers considering the sub-categories introduced in Section~\ref{sec_sota}.}
    \label{fig_chart_categories}
\end{figure}

Moreover, in a more detailed chart, Fig~\ref{fig_chart_categories} shows the distribution of papers surveyed regarding the introduced sub-categories.
It can be seen that most of the papers have targeted \textit{image fusion} techniques for nighttime pedestrian detection applications.
Since thermal imaging (\ie long wavelength \ac{IR}) can capture the infrared radiation from objects and are sensitive to temperature changes, thermal images provide clearer contours information of pedestrians under insufficient lighting conditions.
However, the thermal \ac{IR} modality lacks visual details such as texture, color, and precise edges of the objects, which can be captured by RGB sensors.
In addition, the quality of visible images is significantly degraded under severe weather conditions, low resolution, and unfavorable lighting.
Considering the characteristics of both visible and thermal sensors, cross-spectral fusion has become a promising alternative solution for overcoming the limitations of an unimodal approach to adapt to the all-weather and all-day situations.
By fusing complementary visual features from multiple modalities, the stability, reliability, and perceptibility of the pedestrian detectors are enhanced.
Despite the great progress made in multi-spectral pedestrian detection, there still exists a large gap between the current artificial vision systems and human vision ability.
Among them, \textit{halfway fusion} covers most of the works, and \textit{late fusion} is the second preferred approach in \textit{image fusion} subcategory.
\textit{Knowledge transfer} and \textit{image specification} methodologies are the following trendy solutions according to the stats in the figure.
It can also be seen that \textit{multi-task} methodologies have not absorbed massive attention among the papers published in recent years in the domain.

\begin{figure}[t]
    \centering
    \includegraphics[width=0.9\columnwidth]{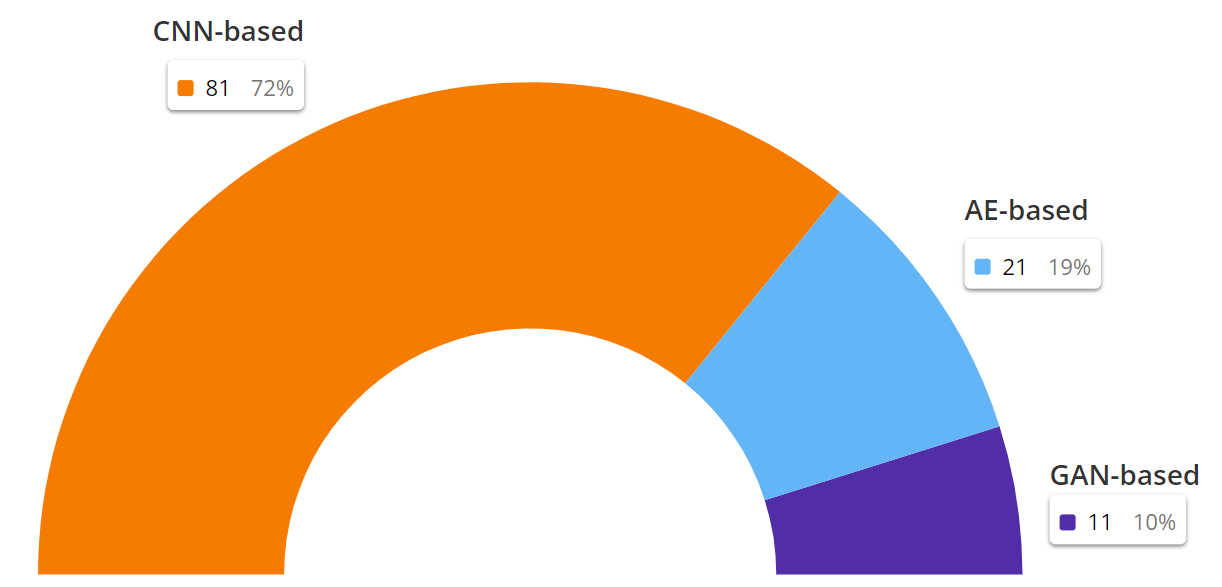}
    \caption{Distribution of deep learning-based architectures for multi-spectral pedestrian detection.}
    \label{fig_chart_dl_apps}
\end{figure}

It should also be added that three types of deep learning-based architectures have been dedicated to achieving multi-spectral pedestrian detection, which can be categorized into the conventional \ac{CNN}-based, Auto-Encoder (AE)-based, and \ac{GAN}-based architectures.
Fig~\ref{fig_chart_dl_apps} shows the distribution of these methodologies and architectures in brief.
Accordingly, end-to-end \ac{CNN}-based methods contain feature extraction, feature fusion, and image reconstruction processes through well-designed loss functions and network architectures.
On the other hand, the AE-based methods first train the encoder and the decoder as the feature extractor and the image reconstructor, respectively.
Then, the multi-image fusion process is accomplished according to the fusion rules.
Finally, and in the \ac{GAN}-based methods, the architecture is suitable for unsupervised pedestrian detection, relying on the adversarial mechanism between the generator and discriminator.
The discriminator forces the generator to make the target distribution in the fused images as close as possible to the source images.

\subsection{Dataset Trends}
\begin{figure}[t]
    \centering
    \includegraphics[width=0.9\columnwidth]{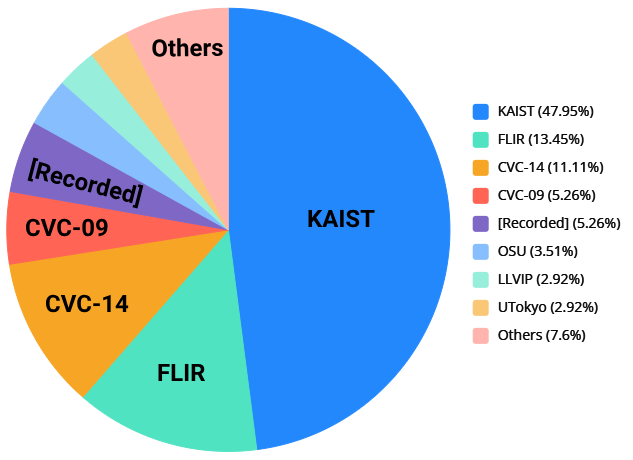}
    \caption{Distribution of the datasets available for evaluation of nighttime pedestrian detection in different works.}
    \label{fig_chart_dataset}
\end{figure}

Regarding the datasets introduced in Section~\ref{sec_baselines}, the surveyed research works have been evaluated on various datasets.
Thus, Fig~\ref{fig_chart_dataset} depicts the distribution of the utilized datasets by the reviewed papers.
It can be seen that most of the papers (\ie around half of them) have utilized \ac{KAIST} dataset.
The second and third datasets other research works use are \ac{FLIR} and CVC-14, respectively.
It should be mentioned that some of the research works (\ie $\sim5.3$ percent) prefer to evaluate their in-house collected, mainly collected from real-world scenarios.
As introduced in Section~\ref{sec_sota}, the differences in the physical characteristics of sensors lead to the misalignment of image pairs and have limited applicability in real-life situations.
Despite the increasing number of visible-\ac{IR} datasets in recent years, accessing instances with strictly aligned multi-spectral images is still a challenging problem.
The benchmark datasets reported in the scientific literature can only provide information under certain scenes, most of which are recorded by a stationary camera.
Therefore, there is a lack of datasets that contain a sufficient variety of fine-grained annotated samples taken from a moving camera, as environments can change dynamically.

\subsection{Performance Evaluations}
Considering the performance of the surveyed works, Table~\ref{tbl_inference_times} analyzes the computational efficiency of the state-of-the-art methods on \ac{KAIST} test set.
It should be noted that the best and second-best results are boldfaced and underlined, respectively.
According to the table, the \ac{MD} \cite{zhang2022low} method tops the chart in processing speed and takes only $0.007$ seconds to process a single image.
The main reason for such performance is due to the theory of knowledge distillation, which accelerates inference by transferring the knowledge learned from a high thermal-resolution model to a low one.
The second best result is the \ac{GAFF} \cite{zhang2021guided} model, which requires only $0.009$ seconds of inference time.
The main reason for such performance is that the \ac{GAFF} only includes three convolution layers, so the total number of learnable parameters and the computational cost is low.
There are some approaches with performances negligibly less than these approaches, including \ac{YOLO}-TGB \cite{vandersteegen2018real} with $0.012$, Dual-\ac{YOLO} \cite{bao2023dual} with $0.016$, \ac{HAFNet} \cite{peng2023hafnet} with $0.017$, and Marnissi \etal \cite{marnissi2022bispectral} with $0.019$ seconds to process a single image.
On the other hand, CMT-CNN \cite{xu2017learning} as a hybrid approach is the most computationally intensive methodology, with $0.59$ seconds to process a single image.
The main reason for such low performance is due to the use of ACF proposals at the test time, which is a time-consuming process.

\begin{table}[t]
    \footnotesize
    \centering
    \caption{Performance evaluation of state-of-the-art deep learning-based multi-spectral pedestrian detectors on \ac{KAIST} test set. The superscripts \textit{X}, \textit{V}, \textit{K}, \textit{P}, \textit{1}, \textit{2}, and \textit{3} represent NVIDIA GPU models used for evaluation, including \textit{TitanX}, \textit{Tesla V100}, \textit{Tesla K40}, \textit{Tesla P40}, \textit{1080Ti}, \textit{2080Ti}, and \textit{3090Ti}, respectively. The best and second-best results are boldfaced and underlined, respectively. Additionally, \textit{s/f} represents seconds per frame.}
    \begin{tabular}{ l l l c }
        \toprule
            \textbf{Family} & \textbf{Backbone} & \textbf{Method} & \textbf{Speed (\textit{s/f})} \\
            \midrule
                DL & CSPDarknet-$53$ & ASPFF Net \cite{fu2021adaptive} & $0.028^{1}$ \\
                & CSPDarknet-$53$ & \ac{MCFF} \cite{cao2021attention} & $0.031^{P}$ \\
                & CSPDarknet-$53$ & \ac{YOLO}-CMN \cite{jiang2022attention} & $0.02^{2}$ \\
                & CSPDarknet-$53$ & Chan \etal \cite{chan2023multispectral} & $0.76^{3}$ \\
                & CSPDarknet-$53$ & \ac{DSMN} \cite{hsia2023all} & $0.76^{3}$ \\
                & Custom \ac{CNN} & \ac{RISNet} \cite{wang2022improving} & $0.1^{V}$ \\
                & Darknet-$19$ & \ac{YOLO}-TGB \cite{vandersteegen2018real} & $0.012^{1}$ \\
                & Darknet-$53$ & TC Det \cite{kieu2020task} & $0.033^{1}$ \\
                & Darknet-$53$ & Marnissi \etal \cite{marnissi2022bispectral} & $0.019^{X}$ \\
                & ELAN & Dual-\ac{YOLO} \cite{bao2023dual} & $0.016^{3}$ \\
                & MobileNet v$2$ & \ac{IT-MN} \cite{zhuang2021illumination} & $0.03^{X}$ \\
                & MobileNet v$3$ & \ac{DMFFNet} \cite{hu2022dmffnet} & $0.021^{2}$ \\
                & ResNet-$18$ & \ac{MD} \cite{zhang2022low} & \textbf{0.007$^1$} \\
                & ResNet-$50$ & Zou \etal \cite{zuo2022improving} & $0.04^{1}$ \\
                & ResNet-$50$ & \ac{MB-Net} \cite{zhou2020improving} & $0.07^{1}$ \\
                & ResNet-$50$ & BAANet \cite{yang2022baanet} & $0.07^{1}$ \\
                & ResNet-$50$ & \ac{HAFNet} \cite{peng2023hafnet} & $0.017^{1}$ \\ 
                & ResNet-$50$ + \ac{FPN} & \ac{ProbEn} \cite{chen2022multimodal} & $0.025^{2}$ \\
                & ResNet-$101$ & ResNet + \ac{FPN} \cite{pei2020fast} & $0.129^{X}$ \\ 
                & VGG-$16$ & \ac{CIAN} \cite{zhang2019cross} & $0.066^{1}$ \\
                & VGG-$16$ & Kim \etal \cite{kim2021uncertainty} & $0.11^{1}$ \\ 
                & VGG-$16$ & \ac{GFD-SSD} \cite{zheng2019gfd} & $0.0512^{1}$  \\
                & VGG-$16$ & \ac{AR-CNN} \cite{zhang2019weakly} & $0.12^{1}$ \\
                & VGG-$16$ & \ac{GAFF} \cite{zhang2021guided} & \underline{$0.009^{1}$} \\
                & VGG-$16$ & \ac{MSR} \cite{kim2022towards}  & $0.04^{1}$ \\                
                & VGG-$16$ & Park \etal \cite{park2018unified} & $0.58^{X}$ \\
                & VGG-$16$ & Halfway Fusion \cite{liu2016multispectral} & $0.43^{X}$ \\
                & VGG-$16$ & IATDNN+IASS \cite{guan2019fusion} & $0.25^{X}$ \\
                & VGG-$16$ & \ac{IAF R-CNN} \cite{li2019illumination} & $0.21^{X}$ \\
                & VGG-$16$ & MSDS-RCNN \cite{li2018multispectral} & $0.22^{X}$ \\
                & VGG-$16$ & LG-FAPF \cite{cao2022locality} & $0.14^{X}$ \\
                & VGG-$16$ & Ding \etal \cite{ding2020convolutional} & $0.222^{X}$ \\
                & VGG-$16$ & HMFFN \cite{cao2019box} & $0.026^{X}$ \\
                & VGG-$16$ & Ding \etal \cite{ding2021robust} & $0.071^{X}$ \\                
                \midrule
                Hybrid & VGG-$16$+ACF & CMT-CNN \cite{xu2017learning} & $0.59^{K}$ \\
        \bottomrule
    \end{tabular}
    \label{tbl_inference_times}
\end{table}

\begin{table*}
    \centering
    \caption{\acf{MR} comparison of state-of-the-art deep learning-based multi-spectral pedestrian detectors in three subsets of the \ac{KAIST} test set, \ie all-day, day-time, and night-time.The best and second-best results are boldfaced and underlined, respectively. Note that the lower \ac{MR} is better.}
    \begin{tabular}{ l l l l c c c }
        \toprule
            \textbf{Method} & \textbf{Family} & \textbf{Category} & \textbf{Backbone} & \textbf{All-Day} & \textbf{Day-Time} & \textbf{Night-Time} \\
            \midrule
                Halfway Fusion \cite{liu2016multispectral} & DL & Halfway-Fusion & VGG-$16$ & $36.99$ & $36.84$ & $35.49$ \\
                Yadav \etal \cite{yadav2020cnn} &  & Halfway-Fusion & VGG-$16$ & $29.00$ & $26.00$ & $32.00$  \\
                \ac{GFD-SSD} \cite{zheng2019gfd} &  & Halfway-Fusion & VGG-$16$ & $28.00$ & $25.80$ & $30.03$  \\ 
                CFR \cite{zhang2020multispectral} &  & Halfway-Fusion & VGG-$16$ & \underline{$6.13$} & $7.68$ & $3.19$  \\
                \ac{CIAN} \cite{zhang2019cross} &  & Halfway-Fusion & VGG-$16$ & $27.71$ & $30.74$ & $21.07$  \\
                Ding \etal \cite{ding2020convolutional} &  & Halfway-Fusion & VGG-$16$ & $34$ & $36$ & $35$  \\
                 \ac{GAFF} \cite{zhang2021guided} &  & Halfway-Fusion & ResNet-$18$ & $7.93$ & $9.79$ & $4.33$  \\
                BAANet \cite{yang2022baanet} &  & Halfway-Fusion & ResNet-$50$ & $7.92$ & $8.37$ & $6.98$  \\
                CS-RCNN \cite{zhang2020attention} &  & Halfway-Fusion & ResNet-$50$ & $11.43$ & $11.86$ & $8.82$  \\
                \ac{HAFNet} \cite{peng2023hafnet} &  & Halfway-Fusion & ResNet-$50$ & $6.93$ & $7.68$ & $5.66$  \\
                Zou \etal \cite{zuo2022improving} &  & Halfway-Fusion & ResNet-$50$ & $7.77$ & $9.41$ & \textbf{2.00}  \\
                ResNet-$101$ + \ac{FPN} + Sum \cite{pei2020fast} &  & Halfway-Fusion & ResNet-$101$ & $27.60$ & $27.92$ & $25.77$  \\
                Yang \etal \cite{yang2023cascaded} &  & Halfway-Fusion & ResNet-$101$ & $10.71$ & $13.09$ & $8.45$  \\ 
                \ac{YOLO}-CMN \cite{jiang2022attention} &  & Halfway-Fusion & CSPDarknet-$53$ & $7.85$ & $8.03$ & $7.82$  \\
                \ac{MCFF} \cite{cao2021attention} &  & Halfway-Fusion & CSPDarknet-$53$ & \textbf{4.91} & \textbf{6.23} & \underline{$2.90$}  \\
                ASPFF Net \cite{fu2021adaptive} &  & Halfway-Fusion & CSPDarknet-$53$ & $11.64$ & $14.14$ & $6.73$  \\
                \ac{DMFFNet} \cite{hu2022dmffnet} &  & Halfway-Fusion & MobileNet v$3$ & $9.26$ & $12.79$ & $5.17$  \\
                \ac{RISNet} \cite{wang2022improving} &  & Halfway-Fusion & Custom CNN & $7.89$ & \underline{$7.61$} & $7.08$  \\
            \midrule
                \ac{IAF R-CNN} \cite{li2019illumination} & DL & Late-Fusion & VGG-$16$ & $15.73$ & $14.55$ & $18.26$  \\
                Ding \etal \cite{ding2021robust} &  & Late-Fusion & VGG-$16$ & $32$ & $34$ & $34$  \\
                LG-FAPF \cite{cao2022locality} &  & Late-Fusion & VGG-$16$ & \textbf{5.12} & \textbf{5.83} & \underline{$3.69$}  \\
                Park \etal \cite{park2018unified} &  & Late-Fusion & VGG-$16$ & $31.36$ & $31.79$ & $30.82$  \\
                \ac{ProbEn} \cite{chen2022multimodal} &  & Late-Fusion & ResNet-$50$+\ac{FPN} & $7.66$ & $9.07$ & $4.89$  \\
                \ac{MS-DETR} \cite{xing2023multispectral} &  & Late-Fusion & ResNet-$50$+ResNet-$18$ & \underline{$6.13$} & \underline{$7.78$} & \textbf{3.18}  \\
                \ac{DSMN} \cite{hsia2023all} &  & Late-Fusion & CSPDarknet-$53$ & $14.33$ & $13.34$ & $22.36$  \\
                \ac{IT-MN} \cite{zhuang2021illumination} &  & Late-Fusion & MobileNet v$2$ & $14.19$ & $14.30$ & $13.98$  \\ 
            \midrule
                Das \etal \cite{das2023revisiting} & DL & Modality-Imbalance & PVT & \textbf{7.41} & \textbf{7.69} & \textbf{7.03} \\ 
                Dasgupta \etal \cite{dasgupta2022spatio} & & Modality-Imbalance & ResNeXt-$50$ & $9.23$ & $9.33$ & $8.97$  \\
                \ac{MB-Net} \cite{zhou2020improving} &  & Modality-Imbalance & ResNet-$50$ & \underline{$8.13$} & \underline{$8.28$} & \underline{$7.86$}  \\
            \midrule
                Kim \etal \cite{kim2019unpaired} & DL & Position-Shift & ResNet-$50$ & $42.89$ & $42.42$ & $43.65$  \\
                AR-CNN \cite{zhang2019weakly} &  & Position-Shift & VGG-$16$ & \underline{$9.34$} & \underline{$9.94$} & \underline{$8.38$}  \\
            Kim \etal \cite{kim2021uncertainty} &  & Position-Shift & VGG-$16$ & \textbf{8.45} & \textbf{9.39} & \textbf{7.39}  \\
                Wanchaitanawong \etal \cite{wanchaitanawong2021multi} &  & Position-Shift & VGG-$16$ & $9.67$ & $10.69$ & $9.24$  \\
            \midrule
                VGG-$16$-two-stage \cite{guo2019domain} & DL & Domain-Adaptation & VGG-$16$ & $46.30$ & $53.37$ & $31.63$  \\
                BU (VLT, T) \cite{kieu2021bottom} &  & Domain-Adaptation & Darknet-$53$ & \textbf{25.61} & \textbf{32.69} & \underline{$10.87$}  \\
                TC-Det \cite{kieu2020task} &  & Domain-Adaptation & Darknet-$53$ & \underline{$27.11$} & \underline{$34.81$} & \textbf{10.31} \\
            \midrule
                Marnissi \etal \cite{marnissi2022unsupervised} & DL & Unsupervised Domain-Adaptation & ResNet-$101$ & $44.60$ & $50.29$ & $28.79$  \\
                U-\ac{TS-RPN} \cite{cao2019pedestrian} &  & Unsupervised Domain-Adaptation & VGG-$16$ & $36.42$ & $37.15$ & $33.00$  \\
                UTL \cite{lyu2022unsupervised} &  & Unsupervised Domain-Adaptation & VGG-$16$ & \textbf{19.98} & \textbf{22.17} & \textbf{15.78}  \\
                Feature-Map Fusion \cite{lyu2021visible} &  & Unsupervised Domain-Adaptation & VGG-$16$ & \underline{$23.09$} & \underline{$24.55$} & \underline{$17.74$}  \\
            \midrule
                Kim \etal \cite{kim2021robust} & DL & Memory-Network & VGG-$16$ & $19.16$ & $24.70$ & $8.26$  \\
                \ac{MSR} \cite{kim2022towards} &  & Memory-Network & ResNet-$101$ & \textbf{10.32} & \textbf{13.28} & \textbf{6.23}  \\
              \midrule 
                IATDNN+IASS \cite{guan2019fusion} & DL & Multi-Task & VGG-$16$ & $26.37$ & $27.29$ & $24.41$  \\
                MSDS-RCNN \cite{li2018multispectral} &  & Multi-Task & VGG-$16$ & \textbf{11.63} & \textbf{10.60} & \textbf{13.73}  \\
            \midrule
                DCRL-PDN \cite{liu2021deep} & DL & Knowledge-Distillation & VGG-$16$ & $25.89$ & $27.01$ & $23.82$  \\
                \ac{MD} \cite{zhang2022low} &  & Knowledge-Distillation & ResNet-$18$ & \textbf{8.03} & \textbf{9.85} & \textbf{4.84} \\
            \midrule
                \ac{LS-GAN} \cite{kieu2021robust} & DL & \ac{I2I}-Translation & Darknet-$53$ & $25.62$ & $31.86$ & $12.92$  \\
            \midrule
                \ac{YOLO}-TGB \cite{vandersteegen2018real} & DL & Transfer-Learning & Darknet-$19$ & $31.2$ & $34.7$ & $23.1$  \\
            \midrule
                Ghose \etal \cite{ghose2019pedestrian} & DL & Saliency-Maps & VGG-$16$ & - & $30.4$ & $21.0$  \\
            \midrule
                Song \etal \cite{song2020full} & DL & Other & ResNet-$50$ & - & $12.23$ & $4.56$  \\
            \midrule
                CMT-CNN \cite{xu2017learning} & Hybrid & - & VGG-$16$+ACF & $49.55$ & $47.30$ & $54.78$  \\
                Choi \etal \cite{choi2016multi} &  & - & VGG-$16$+ACF & $47.31$ & $49.31$ & \underline{$43.75$}  \\
                Kim \etal \cite{kim2018pedestrian} &  & - & VGG-$16$ & \underline{$45.36$} & \textbf{41.30} & $55.82$  \\
                Chen \etal \cite{chen2021robust} &  & - & Darknet-$53$ & \textbf{43.25} & \underline{$46.99$} & \textbf{35.84}  \\
        \bottomrule
    \end{tabular}
    \label{tbl_miss_rates}
\end{table*}

Moreover, Table~\ref{tbl_miss_rates} shows the detection accuracy in evaluating different approaches.
The results are reported in terms of \ac{MR} under \textit{Reasonable} settings, and the approaches are classified according to the categories presented in Section~\ref{sec_sota}.
As shown in the Table, the \ac{MCFF} \cite{cao2021attention} ranks first as a halfway-fusion strategy in overall performance on the \ac{KAIST} by a large margin.
The main reason for such performance is due to the \ac{MCFF} transferring the fusion information from the bottom to the top at different stages.
It can be observed that in the \textit{Reasonable} nighttime criteria, the \ac{MCFF} \cite{cao2021attention} obtains superior results than its daytime experiment.
The main reason for such performance is that the \ac{MCFF} uses the illumination information to learn the fusion weights.
Similarly, LG-FAPF \cite{cao2022locality} as a late-fusion strategy performs remarkably better compared to the other detectors.
The main reason for such performance is due to a locality-guided pixel-level fusion scheme that aggregated the human-related features in the complementary modalities to integrate the prediction confidence scores in color and thermal channels.
Among these methods, only four methodologies (\ie CMT-CNN \cite{xu2017learning}, Kim \etal \cite{kim2018pedestrian}, Choi \etal \cite{choi2016multi}, and Chen \etal \cite{chen2021robust}) are hybrid approaches, which witnessed a significant drop in \ac{MR}.
It can be concluded that the hybrid approaches are not applicable to around-the-clock applications, and specifications are required.

As the final discussion, it is essential to note that by expanding the use of fully autonomous vehicles, the challenges of correct and real-time detecting pedestrians under various scenarios are becoming inevitable.
Accordingly, explainable and interpretable mechanisms to exploit why a system failed/succeeded in a scenario can bring about more public reliability and confidence among people, including pedestrians, for interacting with autonomous systems.
Thus, tailoring the current methodologies with the field of Explainable AI (xAI) is another direction to be investigated by researchers.
\section{Conclusions}
\label{sec_conclude}

The paper in hand provided a comprehensive survey of pedestrian detection approaches tailored to low-light conditions, addressing a crucial challenge in computer vision, surveillance, and autonomous driving.
The accurate and reliable recognition and tracking of pedestrians under reduced visibility is of paramount importance for enhancing the safety of autonomous vehicles and preventing accidents.
The survey has examined a wide array of methodologies, including deep learning-based, feature-based, and hybrid approaches, which have demonstrated promising results in improving pedestrian detection performance in challenging lighting scenarios.
By delving into the current landscape of low-light pedestrian detection, this work contributes to advancing more secure and dependable autonomous driving systems and other applications related to pedestrian safety.
It has also identified ongoing research directions in the field and highlighted potential zones that warrant further research and investigation.
The insights provided in this paper aim to inform and inspire future work, ultimately driving innovation and progress in the domain of pedestrian detection under adverse conditions.

\bibliographystyle{IEEEtran}
\bibliography{references}

\end{document}